\documentclass[11pt]{article}

\usepackage{tikz}
\usepackage{multicol}
\usetikzlibrary{positioning}
\usetikzlibrary{decorations.pathreplacing}

\usepackage[colorlinks = true,
            linkcolor = blue,
            urlcolor  = blue,
            citecolor = blue,
            anchorcolor = blue]{hyperref}
\usepackage{amsmath}
\usepackage{amsthm}
\usepackage{amssymb}
\usepackage{xcolor}
\usepackage{algorithm,caption}
\usepackage{cleveref}
\usepackage{graphicx}
\usepackage{scalerel}
\usepackage{enumitem}
\usepackage{natbib}
\usepackage{framed}
\usepackage[most]{tcolorbox}

\usepackage{mathtools}

\usepackage{booktabs}
\usepackage{tabularx}

\newtcolorbox{redbox}{
    arc=0pt,
    boxrule=0pt,
    colback=red,
    width=.8\textwidth,   
    colupper=black,
    fontupper=\bfseries
}

\usepackage{algpseudocode}

\definecolor{gray}{gray}{0.4}

\definecolor{captioncolor}{RGB}{128,128,128}

\theoremstyle{plain}
\newtheorem{theorem}{Theorem}[section]
\newtheorem{lemma}[theorem]{Lemma}

\newtheorem{prop}[theorem]{Proposition}

\newtheorem{corollary}[theorem]{Corollary}
\newtheorem{question}[theorem]{Problem}
\newtheorem{question*}{Question}
\newtheorem{conjecture}[theorem]{Conjecture}

\newtheorem{observation}[theorem]{Observation}
\newtheorem{assumption}[theorem]{Assumption}

\crefname{claim}{Claim}{Claims}
\crefname{fact}{Fact}{Facts}

\theoremstyle{definition}

\newtheorem{definition}[theorem]{Definition}
\newtheorem{remark}[theorem]{Remark}
\newtheorem{example}[theorem]{Example}
\newtheorem{goal}[theorem]{Goal}
\newtheorem{goal*}{Goal}

\theoremstyle{plain}

\newtheorem{innerrestatedtheorem}{Theorem}

\newenvironment{rtheorem}[1]
  {%
    \renewcommand{\theinnerrestatedtheorem}{\ref*{#1}}%
    \begin{innerrestatedtheorem}[Restated]%
  }
  {%
    \end{innerrestatedtheorem}%
  }

\newtheorem{innerrestatedlemma}{Lemma}

\newenvironment{rlemma}[1]
  {%
    \renewcommand{\theinnerrestatedlemma}{\ref*{#1}}%
    \begin{innerrestatedlemma}[Restated]%
  }
  {%
    \end{innerrestatedlemma}%
  }

\newenvironment{rdefinition}[1]{\renewcommand\thedefinition{\ref*{#1}}\begin{definition}[Restated]}{\end{definition}}
\tcbuselibrary{theorems}

\newtcbtheorem[use counter=theorem, number within=section]{boxtheorem}{Theorem}%
{colback=black!4,
 colframe=black!30,
 coltitle=black,
 colbacktitle=black!4,
 fonttitle=\bfseries,
 separator sign=:,
 enhanced,
 sharp corners,
 boxrule=0.4pt,
 boxed title style={size=small}}{Thm}

\newtcolorbox{technicalresult}{
 enhanced,
 breakable,
 sharp corners,
 frame hidden,
 boxrule=0pt,
 colback=black!1,
 borderline west={0.65pt}{0pt}{black!25},
 left=8pt,right=8pt,top=4pt,bottom=4pt,
 before skip=9pt,after skip=9pt,
 before upper={\setlength{\topsep}{0pt}}
}

\newtcbtheorem[ number within=section]{boxdef}{Definition}%
{colback=definitioncolor,
 colframe=black,
 coltitle=black,
 colbacktitle=definitioncolor,
 fonttitle=\bfseries,
 separator sign=:,
 enhanced,
 boxed title style={size=small}}{Def}

\definecolor{ama-iro}{RGB}{0, 158, 243.0}
\definecolor{fuyu-gaki}{RGB}{251, 74, 52}
\definecolor{momiji}{RGB}{245, 70, 111}
\definecolor{hotaru-bi}{RGB}{229,221,58} 
\definecolor{kon-peki}{RGB}{1,120,217}
\definecolor{shin-kai}{RGB}{77,98,152}
\definecolor{shin-ryoku}{RGB}{1,145,97}
\definecolor{yama-budo}{RGB}{171,14,122}

\definecolor{citecolor}{HTML}{5E9100}

\definecolor{theoremcolor}{HTML}{FFE1D9}
\definecolor{resultcolor}{HTML}{FCDFBE}
\definecolor{lemmacolor}{HTML}{E9E9E9}

\definecolor{constraintcolor}{HTML}{BBD49B}
\definecolor{goalcolor}{HTML}{CAE4A7}

\definecolor{remarkcolor}{HTML}{E3EEC7}

\definecolor{definitioncolor}{HTML}{FFFDDA}
\definecolor{examplecolor}{HTML}{F9E9D9}
\definecolor{questioncolor}{HTML}{DDE8EB}
\definecolor{conjecturecolor}{HTML}{DDE8EB}

\definecolor{lightblue}{HTML}{C3E1FF}

\definecolor{captioncolor}{RGB}{128,128,128}

\tcolorboxenvironment{boxlemma}{colback=lemmacolor, colframe=black,
    colbacktitle=lemmacolor, coltitle=lemmacolor}

\tcolorboxenvironment{rboxlemma}{colback=resultcolor, colframe=black,
	colbacktitle=resultcolor, coltitle=resultcolor}

\tcolorboxenvironment{boxproposition}{colback=resultcolor, colframe=black,
    colbacktitle=resultcolor, coltitle=resultcolor}

\tcolorboxenvironment{boxcorollary}{colback=resultcolor, colframe=black,
    colbacktitle=resultcolor, coltitle=resultcolor}

\tcolorboxenvironment{boxobservation}{colback=resultcolor, colframe=black,
    colbacktitle=resultcolor, coltitle=resultcolor}

\tcolorboxenvironment{boxquestion}{colback=questioncolor, colframe=black,
    colbacktitle=questioncolor, coltitle=shin-kai}
\newenvironment{boxquestion*}{\begin{question*}}{\end{question*}}
\tcolorboxenvironment{boxquestion*}{colback=questioncolor, colframe=black,
    colbacktitle=questioncolor, coltitle=questioncolor}

\tcolorboxenvironment{boxdefinition}{colback=definitioncolor, colframe=black,
    colbacktitle=definitioncolor, coltitle=definitioncolor}

\tcolorboxenvironment{rboxdefinition}{colback=definitioncolor, colframe=black,
	colbacktitle=definitioncolor, coltitle=definitioncolor}

\tcolorboxenvironment{boxassumption}{colback=definitioncolor, colframe=black,
    colbacktitle=definitioncolor, coltitle=definitioncolor}

\tcolorboxenvironment{boxexample}{colback=examplecolor, colframe=black,
    colbacktitle=examplecolor, coltitle=examplecolor}

\newtheorem{exercise}{Exercise}

\tcolorboxenvironment{boxexercise}{colback=examplecolor, colframe=black,
    colbacktitle=examplecolor, coltitle=examplecolor}

\tcolorboxenvironment{boxgoal}{colback=goalcolor, colframe=black,
    colbacktitle=goalcolor, coltitle=goalcolor}

\newenvironment{boxgoal*}{\begin{goal*}}{\end{goal*}}
\tcolorboxenvironment{boxgoal*}{colback=goalcolor, colframe=black,
    colbacktitle=goalcolor, coltitle=goalcolor}

\tcolorboxenvironment{boxremark}{colback=remarkcolor, colframe=black,
    colbacktitle=remarkcolor, coltitle=remarkcolor}

\tcolorboxenvironment{boxconjecture}{colback=conjecturecolor, colframe=black,
    colbacktitle=conjecturecolor, coltitle=shin-kai}
\newenvironment{boxconjecture*}{\begin{conjecture*}}{\end{conjecture*}}
\tcolorboxenvironment{boxconjecture*}{colback=conjecturecolor, colframe=black,
    colbacktitle=conjecturecolor, coltitle=conjecturecolor}

\usepackage{geometry}
\newtheorem{innercustomthm}{Theorem}
\newenvironment{customthm}[1]
  {\renewcommand\theinnercustomthm{#1}\innercustomthm}
  {\endinnercustomthm}

\newcommand{\case}[4]{ \begin{cases}
    #1\quad &#2\\
    #3\quad &#4\end{cases}}
    \newcommand{\inP}[2]{\langle #1,#2\rangle}

\newcommand{\VC}{\mathtt{VC}}
\newcommand{\R}{\mathbb{R}}

\newcommand{\W}{\mathcal{W}}

\newcommand{\X}{\mathcal{X}}

\renewcommand{\H}{\mathcal{H}}
\newcommand{\F}{\mathcal{F}}
\newcommand{\G}{\mathcal{G}}
\newcommand{\B}{\mathcal{B}}

\newcommand{\eps}{\varepsilon}

\newcommand{\supp}{\mathsf{supp}}

\newcommand{\D}{\mathcal{D}}

\newcommand{\U}{\mathcal{U}}

\newcommand{\Ex}{\mathop{\mathbb{E}}}
\renewcommand{\Pr}{\mathop{\mathbb{P}}}

\newcommand{\Maj}{\mathtt{Maj}}
\newcommand{\Rep}{\mathtt{Rep}}
\newcommand{\Rad}{\mathtt{Rad}}

\newcommand{\fat}{\mathtt{fat}}

\newcommand{\norm}[1]{\left\lVert#1\right\rVert}

        \hypersetup{
           breaklinks=true,   
           colorlinks=true,   
        }

\author{
Noga Alon\thanks{Department of Mathematics, Princeton University, Princeton, NJ, USA. Research supported in part by NSF grant DMS-2553988 and by
AFOSR grant FA9550261B300.},
Steve Hanneke\thanks{Department of Computer Science, Purdue University, West Lafayette, IN, USA. Research supported by grant no.\ 2024243 from the United States - Israel Binational Science Foundation (BSF).},
Shay Moran\thanks{Departments of Mathematics, Computer Science, and Data and Decision Sciences, Technion--Israel Institute of Technology; Google Research.\ Shay Moran acknowledges support by Israel by ISF grant 2362/26. and by the European Union (ERC, GENERALIZATION, 101039692).},
and Tom Waknine\thanks{Department of Mathematics, Technion--Israel Institute of Technology, Haifa, Israel.}
}

\title{Uniform convergence at optimal rate and scale }
\begin{document}
\maketitle

\begin{abstract}
Uniform convergence asks how many random samples suffice for empirical
averages to approximate expectations simultaneously over an entire class
of functions. It is a central concept in probability and a foundation of statistical learning. For binary-valued classes, the Vapnik--Chervonenkis theorem
characterizes uniform convergence by finiteness of the VC dimension. For dimension $d$, the sharp worst-case sample
complexity for additive error $\varepsilon$ is of order~$d/\varepsilon^2$. For real-valued functions, the analogous combinatorial parameter is the fat-shattering dimension, which depends on an approximation scale~$\gamma$. The challenge is to achieve error $\gamma+\varepsilon$ with order
$d/\varepsilon^2$ samples, matching the sample bound for binary
classes, where $d$ is now the fat-shattering dimension at scale $\gamma$. Previous results achieved either the optimal scale or the optimal
sample complexity, but not both. Our main result closes this gap,
establishing the sharp sample complexity at the optimal scale.

Along the way, we use several tools that may be of independent interest. The first
concerns $\varepsilon$-nets with margin. A classical $\varepsilon$-net
intersects every set of mass at least $\varepsilon$ in a prescribed
family; we allow the net to intersect a slightly enlarged version of
each such set. Remarkably, this relaxation permits bounds independent
of the ambient dimension in natural geometric settings. Second, we derive improved covering bounds in both discrete and real-valued settings: we extend Haussler’s covering lemma to partial concept classes and use this extension to obtain \(\ell_2\) covers at the optimal scale for real-valued function classes. Third, we obtain a chaining bound at the optimal scale by aligning the covers so that successive approximations become increasingly close to one another. We also highlight open problems and directions for future research arising from these results.

\end{abstract}

\section{Introduction}

Suppose we wish to estimate the expectation of a bounded function $f$ under an unknown distribution. We can draw independent samples and average the values of $f$ on these samples. The Law of Large Numbers guarantees that this average converges to the expectation. Hoeffding's Inequality \cite{hoeffding1963probability} makes this quantitative: $O(\varepsilon^{-2}\log(1/\delta))$ samples suffice for error at most $\varepsilon$ with confidence $1-\delta$. 
Suppose now we wish to estimate the expectations of $M$ such functions. A union bound gives the simultaneous guarantee with
\[
O\left(\frac{\log M+\log(1/\delta)}{\varepsilon^2}\right)
\]
samples. Thus, the number of functions enters only logarithmically.

Uniform convergence asks when a single sample can accurately estimate all expectations in an infinite class~$\F$ of functions\footnote{Throughout the paper, we assume that $
\F$ satisfies some mild measurability conditions such as \cite[Appendix C]{Pollard1984ConvergenceOS}}. The required sample size may depend on $\F$, the desired accuracy, and the confidence, but must work for every underlying distribution. Classes $\F$ for which the empirical averages uniformly approximate the expectations are called \emph{uniform Glivenko--Cantelli classes} \cite{dudley1991uniform}.

A classical example is estimating a cumulative distribution function
(CDF). For independent samples $S=(x_1,\ldots,x_n)$ from a distribution
$\D$ on $\R$, the true and empirical CDFs are
\[
F_\D(t)=\Pr_{x\sim\D}[x\leq t],
\qquad
\widehat F_S(t)=\frac1n\sum_{i=1}^n\mathbf{1}_{\{x_i\leq t\}}.
\]
These two CDFs are the expectations and sample averages of the
threshold functions $f_t(x)=\mathbf{1}_{\{x\leq t\}}$. Uniform convergence
for this class means that the same sample estimates $F_\D(t)$ accurately
for all thresholds simultaneously. The Glivenko--Cantelli Theorem,
proved in 1933 \cite{glivenko1933empirica,cantelli1933empirica},
guarantees this accuracy with high probability for sufficiently large
samples. An entire CDF can thus be estimated without a parametric model.
The theorem is therefore sometimes called the
\emph{Fundamental Theorem of Mathematical Statistics}
\cite[p.400]{renyi2007probability}.

\paragraph{Binary-valued functions.}
For classes of binary-valued functions, Vapnik and Chervonenkis
\cite{Vapnik1971ChervonenkisOT} showed that uniform convergence holds
precisely when the VC dimension is finite. This characterization also
forms a major part of the Fundamental Theorem of PAC Learning \cite{VC-lrn}.
Their proof also bounds the number of samples needed. For a class of
VC dimension $d$, their classical symmetrization argument, based on an
independent \emph{ghost sample}, gives
\[
m(\varepsilon,\delta)
=
O\left(
\frac{d\log(1/\varepsilon)+\log(1/\delta)}{\varepsilon^2}
\right),
\]
where $\varepsilon$ is the approximation error and $\delta$ is the
allowed failure probability.

The extra logarithmic factor can be removed by combining \emph{chaining}
with bounds on the \emph{covering numbers} of the function class.
Dudley's chaining method \cite{dudley1978central},
Haussler's covering estimates \cite{haussler1995sphere}, and Talagrand's
bounds for empirical processes \cite{talagrand1994sharper} yield the
optimal worst-case sample complexity
\[
\Theta\left(\frac{d+\log(1/\delta)}{\varepsilon^2}\right),
\qquad d=\VC(\F).
\]

This gives a tight theorem for binary classes, using techniques with applications in probability, functional analysis, and theoretical computer science. Our proof refines the covering and chaining tools, which we recall in Section~\ref{Sec:Technical-Contributions}.


\paragraph{Real-valued functions.}
The binary case corresponds to estimating probabilities of events.
More generally, we often want to estimate expectations of real-valued
functions. For such classes, VC dimension is replaced by a
scale-dependent extension, called the \emph{fat-shattering dimension}
\cite{KEARNS1994}.

Fix a scale $\gamma>0$. The dimension $\fat_\gamma(\F)$ is the largest
$n$ for which there are points $x_1,\ldots,x_n$ and thresholds
$r_1,\ldots,r_n$ with the following property: for each of the $2^n$ ways to label the points {$x_1,\ldots,x_n$} \emph{high} or
\emph{low}, there is a function $f\in\F$ satisfying all the
corresponding inequalities simultaneously:
\[
\begin{aligned}
f(x_i)&\geq r_i+\frac{\gamma}{2}
&&\text{if $x_i$ is labeled \emph{high}},\\
f(x_i)&\leq r_i-\frac{\gamma}{2}
&&\text{if $x_i$ is labeled \emph{low}}.
\end{aligned}
\]
The thresholds are chosen once, while $f$ may depend on the labeling.


For $\{0,1\}$-valued classes and $\gamma=1$, this recovers VC dimension:
realizing these patterns is exactly realizing every binary labeling. For real-valued classes, the fat-shattering dimension may decrease as the scale increases. Classical work showed that a bounded class satisfies uniform convergence exactly when its fat-shattering dimension is finite at every positive scale \cite{Alon1997-scale-sensitive,BARTLETT1998}.
These classical works reduce the study of uniform convergence to
understanding what can be guaranteed at each scale:
\begin{center}
\emph{If the fat-shattering dimension is finite at a
scale~$\gamma$, how accurately can we estimate the expectations
of all functions in the class simultaneously?} 
\end{center}
Without information at finer scales, we cannot in general guarantee
error below $\gamma$. The natural target is therefore
$\gamma+\varepsilon$, where the additional error $\varepsilon$ can
be reduced by drawing more samples. Our goal is to achieve this with
the same sample complexity as in the binary case. We denote the
required sample size, at confidence $1-\delta$, by
$m_\gamma(\varepsilon,\delta)$. If this sample size is finite for every
accuracy and confidence, the class satisfies
\emph{$\gamma$-uniform convergence}.

The classical bounds gave error $C\gamma+\varepsilon$, for a universal constant $C>1$, with additional logarithmic factors in the sample complexity \cite{Alon1997-scale-sensitive,BARTLETT1998}. Colomboni, Esposito, and Paudice \cite{COLOMBONI2025-improved-UC}, using bounds of Rudelson and Vershynin \cite{rudelson2006combinatorics}, obtained the optimal binary rate, but with error $C\gamma+\varepsilon$ for a universal constant $C>1$. The result of \cite{aiyer2026scale} reaches error $\gamma+\varepsilon$, but retains additional logarithmic factors in the sample complexity. Thus, the optimal rate and the optimal scale were available separately.

\paragraph{Our contribution.}
We obtain both simultaneously:\par\nopagebreak[4]
\begin{boxtheorem}{Main result}{Main result}

Let $\F\subseteq[-1,+1]^\X$, let $\gamma\in(0,1)$, and suppose that $d:=\fat_\gamma(\F)<\infty$. Then $\F$ satisfies $\gamma$-uniform convergence, with
\[
 m_\gamma(\varepsilon,\delta)
 =O\left(\frac{d+\log(1/\delta)}{\varepsilon^2}\right)
 \qquad \eps,\delta\in(0,1).
\]
Moreover, for every $\gamma\in(0,1)$ and integer $d\geq1$, there is a class $\F$ with $\fat_\gamma(\F)=d$ such that, for all sufficiently small $\varepsilon,\delta>0$,
\[
 m_\gamma(\varepsilon,\delta)
 =\Omega\left(\frac{d+\log(1/\delta)}{\varepsilon^2}\right).
\]
\end{boxtheorem}

The theorem gives the optimal binary sample complexity at error $\gamma+\varepsilon$. The matching lower bound shows that this cannot be improved in general using only the fat-shattering dimension at scale~$\gamma$. Particular classes may admit stronger guarantees when more is known about their behavior at finer scales. Our proof combines covering bounds with chaining at optimal scale; the full \hyperref[Proof:Main result]{proof} appears in Section~\ref{Sec:Main-Result-Proof}.

In section \ref{Sec:Technical-Contributions} we give an self-contained presentation overview of technical tools used to prove our main result. In section \ref{Sec:Proof-Over} we give a proof overview of the main results in the paper. And finally, in section \ref{Sec:Technical-Proofs} we give formal proofs for all results in the paper.

\section{Technical contributions}\label{Sec:Technical-Contributions}

Before turning to the proof of our uniform-convergence bound, we give a self-contained presentation of the technical tools behind it. Our aim is to explain the ideas in their natural settings and relate them to earlier work, so that the techniques can be understood beyond the present proof. This also brings out results that may be of independent interest, including some that are not used directly in our proof but illustrate the techniques more broadly.

We begin with $\varepsilon$-nets with margin and rounding fractional covers, then present covering bounds for partial binary classes and real-valued classes, and conclude with chaining at optimal scale in Euclidean space. The subsections can be read largely independently. In Section~\ref{Sec:Proof-Over}, we return to uniform convergence and explain how these tools combine to prove our main result.

\subsection{\texorpdfstring{$\varepsilon$}{ε}-nets with margin and rounding fractional covers}\label{Sec:Margin-Nets}

A basic question in combinatorics is how to choose a small set of points that intersects every sufficiently large set in a given family. Let $\X$ be a domain, let $X\subset\X$ be a set of $n$ points, and let $\U$ be a family of subsets of $\X$. A set $E\subseteq\X$ is an $\varepsilon$-net for $\U$ with respect to $X$ if it intersects every $U\in\U$ with $\lvert U\cap X\rvert\geq\varepsilon n$. The Haussler--Welzl theorem \cite{haussler1986net} guarantees that
\[
O\left(\frac{d}{\varepsilon}\log\frac{1}{\varepsilon}\right)
\]
points sampled independently and uniformly from $X$ form such a net with constant probability, where~$d$ is the VC dimension of the set family $\U$. Importantly, this bound is independent of $n$.

\paragraph{Halfspaces with margin.}
Halfspaces in $\R^d$ provide a classical illustration of this theorem. Their VC dimension is $d+1$, so the Haussler--Welzl theorem \cite{haussler1986net} guarantees that, given $n$ points, one can select $O((d/\varepsilon)\log(1/\varepsilon))$ of them to meet every halfspace containing at least $\varepsilon n$ points. Conversely, for $d\geq2$, there are point sets for which any such selection requires $\Omega(d/\varepsilon)$ points.\footnote{In dimension one, halfspaces are rays, and the minimum and maximum of $X$ hit every ray that intersects $X$.}
The logarithmic gap between these upper and lower bounds has attracted considerable attention: Matou{\v s}ek, Seidel, and Welzl removed it for $d=2,3$ \cite{matousek1990net}, while Pach and Tardos showed that it is necessary already in dimension four \cite{pach2011nets-lower}. Kupavskii, Mustafa, and Pach \cite{kupavskii2016nets} subsequently established the matching lower bound $\Omega((d/\varepsilon)\log(1/\varepsilon))$ for all $d\geq4$, settling the dependence on both parameters.\footnote{These lower bounds concern strong $\varepsilon$-nets, whose points must belong to $X$.}

We consider a natural relaxation of $\varepsilon$-nets that allows a \emph{margin}: it suffices to hit each large halfspace after moving its boundary outward by a distance $\eta$. Equivalently, place a ball of radius $\eta$ around each net point and require every large halfspace to intersect one of these balls. 
This margin viewpoint has been studied through partial concept classes \cite{Dense2025Bourneuf,ivanov2026partial-margin}. It also appears in work on no-dimensional variants of Carath\'eodory's, Helly's, and Tverberg's theorems and on weak $\varepsilon$-nets with margin for convex sets; see, e.g., \cite{adiprasito2020withoutdimension,ivanov2021nodimension,choudhary2022nodimensional,harpeled2024tverberg,barabanshchikova2026tight,ivanovkadets2026optimality}.

The following theorem gives a bound independent of the Euclidean dimension. It applies to $\ell_2$ geometry in any real Hilbert space, including infinite-dimensional ones.

\begin{technicalresult}
\begin{theorem}[Halfspace nets with margin in Hilbert space]\label{Thm:Net-Hilbert-Halfspaces}
Let $X$ be a set of $n$ points in the unit ball\footnotemark{} of a real Hilbert space $\X$, and let $\varepsilon,\eta\in(0,1)$. Then, there is a set $E\subseteq\X$ of size $O(1/(\varepsilon\eta^2))$ such that every closed halfspace containing at least $\varepsilon n$ points of $X$ has distance at most $\eta$ from some point of $E$.

The bound is optimal: for all sufficiently small $\varepsilon,\eta>0$, there is a finite set $X$ in the unit ball of~$\ell_2$ for which every such net has $\Omega(1/(\varepsilon\eta^2))$ points.
\end{theorem}
\end{technicalresult}
\footnotetext{For points in a ball of radius $R$, the same bound holds with margin $\eta R$. Rescaling the lower bound in the \hyperref[Proof:Net-Hilbert-Halfspaces]{proof} shows that dependence on the ratio between the radius and the margin is necessary.}

This improves upon a result of Bourneuf, Charbit, and Thomass{\'e} \cite{Dense2025Bourneuf}, who extended the Haussler--Welzl theorem to partial concept classes. 
Combining their extension with the classical bound of order $1/\eta^2$ on the VC dimension of halfspaces with margin $\eta$ gives a net of size $O((\varepsilon\eta^2)^{-1}\log(1/\varepsilon))$ in Hilbert space. Theorem~\ref{Thm:Net-Hilbert-Halfspaces} removes this logarithmic factor and gives the optimal dependence on $\varepsilon$ and $\eta$. Its full \hyperref[Proof:Net-Hilbert-Halfspaces]{proof} appears in Section~\ref{Sec:Technical-Proofs}.

It is interesting to compare this bound with the classical bounds without margin. Our bound is linear in $1/\varepsilon$, whereas the classical bound has an extra factor of $\log(1/\varepsilon)$ for $d\geq4$. Another difference is that our nets are \emph{weak}, or \emph{external}: their points may lie anywhere in the ambient space. They are not obtained by randomly sampling points from $X$, as in the Haussler--Welzl construction. Indeed, independent uniform sampling from $X$ can still require the logarithmic factor even with a margin, as follows from lower bounds for margin-based generalization; see, e.g., \cite[Lemma~11]{gronlund2020margin}.

Beyond the Hilbert-space setting, Ivanov et al.\ also study margins for halfspaces in other normed spaces and for balls in $L_p$ spaces \cite{ivanov2026partial-margin}. It would be interesting to determine the optimal sizes of margin nets in these settings.

\subsubsection{Rounding fractional covers}
It is natural to study $\varepsilon$-nets with margin for other geometric
families, such as polytopes of bounded complexity or even arbitrary convex
sets. In each case, allowing a margin means slightly enlarging the set
by moving its boundary outward. For balls, this simply increases the
radius while keeping the center fixed. As with halfspaces, this relaxation
yields dimension-independent bounds for balls.
While the result holds in more general spaces, we present it in Hamming space. Both because this is the setting
relevant to our proof and because it illustrates the applicability of
margin nets in a more combinatorial setting. We use these nets to obtain a fractional extension of the familiar packing--covering relation.

For a finite set $A$ in a metric space $(Z,\rho)$, let $P_r(A)$ be the maximum number of pairwise disjoint closed balls of radius $r$ centered in $A$, and let $C_r(A)$ be the minimum number of radius-$r$ balls that cover~$A$, with centers anywhere in $Z$. These are the matching and transversal numbers of the ball family $\{B_r(a):a\in A\}$: hitting every ball is equivalent to covering its center. The familiar packing--covering inequality is
\[
    C_{2r}(A)\leq P_r(A)\leq C_r(A).
\]
Indeed, a maximal family of disjoint radius-$r$ balls yields a $2r$-cover by the triangle inequality, while disjoint balls require distinct hitting points.

In a fractional cover, each center receives a nonnegative weight, and the centers within distance $r$ of each $a\in A$ must have total weight at least one. A fractional packing assigns nonnegative weights to the balls $B_r(a)$, with total weight at most one on the balls containing any given point. Linear-programming duality equates the minimum covering weight and maximum packing weight:
\[
    C_r^*(A)=P_r^*(A).
\]
The classical relation therefore gives $C_{2r}(A)\leq P_r^*(A)=C_r^*(A)$.

We show that, in Hamming space, a fractional cover can be rounded with a radius increase of only $1+\eta$, at the expense of more centers. We use the normalized Hamming distance $d_H$ on $\{0,1\}^n$, which is the fraction of coordinates on which two vectors differ, and write $B_r(x)=\{y:d_H(x,y)\leq r\}$. All packing and covering numbers below use the cube as the ambient space.

\begin{technicalresult}

\begin{theorem}[Rounding fractional covers]\label{Thm:Fractional-Cover-Rounding}
Let $A\subseteq\{0,1\}^n$ be nonempty, and let $r,\eta\in(0,1)$. Write $M=C_r^*(A)=P_r^*(A)$ for the common value of the fractional covering and fractional packing numbers at radius $r$. Then
\[
    C_{(1+\eta)r}(A)
    =O\left(\frac{M}{\eta^2}\log(eM)\right).
\]
\end{theorem}
\end{technicalresult}

Thus, for any fixed $\eta>0$, rounding costs only a logarithmic factor relative to the fractional covering weight, while increasing the radius by a factor of $1+\eta$.

The connection to $\varepsilon$-nets is simple. Normalize the weights of a fractional cover of total weight $M$ to a probability distribution $\mu$ on the possible centers. Every ball $B_r(a)$, $a\in A$, then has mass at least $1/M$. A margin net that meets each of these balls after its radius is increased to $(1+\eta)r$ places a point within distance $(1+\eta)r$ of every $a\in A$. Its points therefore form the desired cover. The following theorem, applied with radius $r$, margin $\eta r$, and mass threshold $1/(2M)$, gives the required bound.\footnote{If $(1+\eta)r\geq1$, one ball already covers the entire cube.}

\begin{technicalresult}
\begin{theorem}[Nets for Hamming balls with margin]\label{Thm:Net-Hamming-Ball}
Let $\mu$ be a probability distribution on $\{0,1\}^n$. Fix a radius $\alpha$ and an additive margin $\eta$ satisfying $0<\eta\leq\alpha$ and $\alpha+\eta\leq1$. For every $\varepsilon\in(0,1)$, there is a set $E\subseteq\{0,1\}^n$ of size
\[
    |E|=O\left(\frac{\alpha^2}{\varepsilon\eta^2}\log\frac{e}{\varepsilon}\right)
\]
{such that every closed Hamming ball of radius $\alpha$ and $\mu$-mass at least $\varepsilon$ intersects $E$ after its radius is increased to $\alpha+\eta$.}

The bound is optimal up to the logarithmic factor for small radii. For every sufficiently small $\varepsilon,\alpha>0$ and every $0<\eta\leq\alpha$, there are a dimension $n$ and a distribution $\mu$ for which every such net has $\Omega(\alpha^2/(\varepsilon\eta^2))$ points.
\end{theorem}
\end{technicalresult}
The upper bound is independent of $n$ and depends on the radius and
margin only through their ratio $\alpha/\eta$. The same upper bound
extends, uniformly in the dimension, to balls in $\ell_p$ for every
$p\in[1,2]$, as explained at the end of the
\hyperref[Proof:Net-Hamming-Ball]{proof} in
Section~\ref{Sec:Technical-Proofs}.

A restriction on the radius is necessary for the lower bound: if
$\alpha+\eta\geq 1/2$, the two antipodal points $\mathbf{1}$ and
$-\mathbf{1}$ meet every enlarged ball. For small radii, a factor of
$\log(1/\varepsilon)$ remains between our upper and lower bounds. The
lower-bound construction also extends to balls in $\ell_p$, with the
same dependence on the parameters, when the net is required to be
strong. It does not, however, give a corresponding lower bound for
weak nets.

\subsection{Covering numbers}\label{Sec:Covering-Numbers}

In the preceding subsection, we studied covers in metric spaces. We now apply this viewpoint to classes of functions on a domain of $n$ points. Each function is identified with the vector of its $n$ values, and a cover consists of finitely many representatives that approximate every function in the class. We consider both binary-valued and real-valued functions. For binary functions, distance is the fraction of domain points on which they disagree; for real-valued functions, we use normalized Euclidean distance.

We first establish covering bounds for partial binary functions, which may be undefined on some inputs, and then use them to cover real-valued classes. This link between partial classes and real-valued learning has been exploited in previous work \cite{BARTLETT1998,Long01,aiyer2026scale}.
The covering bounds we present build on the $\varepsilon$-net machinery of Subsection~\ref{Sec:Margin-Nets}. Combined with the chaining result in Subsection~\ref{Sec:Refined-Chaining}, these bounds yield our uniform-convergence guarantee, as explained in Section~\ref{Sec:Proof-Over}.

\subsubsection{Covering partial binary classes}\label{Sec:Partial-Covers}

Consider a class $\F$ of binary-valued functions on a domain of $n$ points. We wish to approximate every function in $\F$, allowing disagreements on at most an $\varepsilon$-fraction of the domain. Haussler's covering lemma \cite{haussler1995sphere} guarantees that this can be done using at most
\[
\left(\frac{C}{\varepsilon}\right)^d
\]
representatives, where $d=\VC(\F)$ and $C$ is a universal constant. Thus, the number of representatives depends only on the VC dimension and the desired accuracy, independently of $n$.

Haussler obtained the optimal dependence on $\varepsilon$, improving an earlier bound of Dudley \cite{dudley1978central} by removing a $\log(1/\varepsilon)$ factor from the base. His result is usually stated as a packing bound; the covering form follows by taking a maximal separated subfamily. Chazelle \cite{chazelle1992packing} simplified the proof, and Kupavskii and Zhivotovskiy \cite{kupavskii2020smallnets} later gave a short proof using the one-inclusion graph algorithm that recovers Haussler's original bound.

We extend this covering bound to functions that may be undefined on some inputs. A \emph{partial binary class} is a family $\F\subseteq\{-1,+1,\star\}^{\X}$, where $\star$ denotes an undefined value. Its VC dimension is the largest size of a set $S\subseteq\X$ on which every \emph{binary} labeling is realized by a member of $\F$ that is defined throughout $S$. When approximating a partial function, we only count disagreements at points where it is defined.
More precisely, for a finite domain $\X$, an \emph{$\varepsilon$-disambiguation} of $\F$ is a class of total binary functions $\H\subseteq\{-1,+1\}^{\X}$ such that, for every $f\in\F$, some $h\in\H$ satisfies
\[
\bigl|\{x\in\X:f(x)\neq\star\text{ and }f(x)\neq h(x)\}\bigr|
\leq\varepsilon|\X|.
\]
Thus, each partial function $f\in\F$ can be completed to some $h\in\H$
after flipping its value on at most an ${\varepsilon}$-fraction of the domain. We prove the following bound.

\begin{technicalresult}
\begin{theorem}[Covering partial binary classes]\label{Thm:Par-Cover}
Let $n\geq 1$. 
Let $\F\subseteq\{-1,+1,\star\}^{n}$ be a partial binary class of VC dimension $d$. For every $\varepsilon\in(0,1/2]$, there is an $\varepsilon$-disambiguation $\H\subseteq\{-1,+1\}^{n}$ satisfying
\[
|\H|\leq(1/\varepsilon)^{cd\log(1/\varepsilon)},
\]
where $c>0$ is a universal constant.

This bound is optimal up to constant factors in the exponent: there are universal constants $c_0,\varepsilon_0>0$ such that, for every integer $d\geq1$ and every $\varepsilon\in(0,\varepsilon_0]$, some partial binary class of VC dimension $d$ has the property that every $\varepsilon$-disambiguation $\H$ satisfies
\[
|\H|\geq(1/\varepsilon)^{c_0d\log(1/\varepsilon)}.
\]
\end{theorem}
\end{technicalresult}

As in Haussler's lemma, the bound is independent of $n$, which will be essential when we cover real-valued classes in Subsection~\ref{Sec:Real-Covers}.

Our bound has an additional logarithmic factor in the exponent. It would be interesting to determine whether the upper bound can be sharpened to $(C/\varepsilon)^{d\log(1/\varepsilon)}$, with a universal constant in the base, as in Haussler's bound for total classes.


Our proof constructs covers directly from learning rules derived from sample compression. These rules give a fractional cover, which the $\varepsilon$-net theorem for partial concept classes \cite{Dense2025Bourneuf} rounds into a finite one. Section~\ref{Sec:Overview-Covers} explains this argument.


As for the lower bound, the case of $d=1$ follows directly from the reduction of Alon, Hanneke, Holzman, and Moran \cite{alon2022theory}, combined with Pabbaraju's improved construction for the Alon--Saks--Seymour problem \cite{pabbaraju2026optimal}. For general $d$, we take the direct product of $d$ copies of the
VC-dimension-one class. Our contribution is to show that the
$\varepsilon$-disambiguation complexity grows multiplicatively under
this construction, up to constant factors in the exponent. We use a
fractional relaxation of exact disambiguation, whose complexity is
exactly multiplicative under direct products by linear-programming
duality. We then transfer the resulting lower bound to integral
$\varepsilon$-disambiguations with only a constant-factor loss in the
exponent.
The full \hyperref[Proof:Par-Cover]{proof} is given in Section \ref{Sec:Technical-Proofs}.


\subsubsection{Covering real-valued classes}\label{Sec:Real-Covers}

We now turn to bounded real-valued functions on a domain of $n$ points, which we identify with a family of vectors $V\subseteq[-1,+1]^n$. As in the binary case, we seek a small collection of representatives that approximates every vector in the family. The relevant complexity measure is the fat-shattering dimension, defined in Section~1.

We measure approximation by the normalized $\ell_2$ distance
\[
\|v-h\|_2=\left(\frac1n\sum_{i=1}^n |v_i-h_i|^2\right)^{1/2}.
\]
We similarly normalize the $\ell_1$ distance by averaging over the coordinates, while the $\ell_\infty$ distance remains the maximum absolute difference. With this normalization, the usual inequalities are reversed:
\[
\|v-h\|_1\leq\|v-h\|_2\leq\|v-h\|_\infty.
\]
Thus, at the same radius, an $\ell_\infty$ cover is also an $\ell_2$ cover, and an $\ell_2$ cover is also an $\ell_1$ cover. Table~\ref{tab:covering-bounds} uses these normalized distances.

A cover of $V$ at radius $r$ is a finite set $H\subseteq\R^n$ such that every $v\in V$ is within distance $r$ of some $h\in H$. The representatives need not belong to $V$. We write $N_2(V,r)$ for the minimum size of such a cover in normalized $\ell_2$ distance, and use $N_1(V,r)$ and $N_\infty(V,r)$ for the other two metrics.

Covering bounds in terms of fat-shattering dimension date back to the 1990s \cite{Alon1997-scale-sensitive,BartlettKulkarniPosner1997,BARTLETT1998}. Table~\ref{tab:covering-bounds} summarizes this development. Two issues arise: how small a covering radius can be obtained from $d=\fat_\gamma(V)$, and whether the number of representatives can be bounded independently of $n$. Classical results give such independent bounds at a radius that is a constant multiple of $\gamma$. Below radius $\gamma/2$, no bound independent of $n$ is possible in general, even when $\fat_\gamma(V)=0$. For example, take the class $(0,\gamma)^n$ of all functions from $[n]$ to $(0,\gamma)$.

More recently, Aiyer et al.\ \cite{aiyer2026scale} obtained covers at radius $\gamma/2+\varepsilon$ in the stronger $\ell_\infty$ metric, which requires approximation within this radius at every coordinate. Their bound depends on $n$, and some dependence on $n$ is unavoidable for $\ell_\infty$ covers. We obtain the same radius in normalized $\ell_2$ distance with a bound independent of $n$.

\begin{table}[t]
\centering
\small
\renewcommand{\arraystretch}{1.25}
\begin{tabularx}{\textwidth}{
    @{}
    >{\raggedright\arraybackslash}p{3.2cm}
    >{\raggedright\arraybackslash}p{1.25cm}
    >{\raggedright\arraybackslash}p{2.2cm}
    X
    c
    @{}
}
\toprule
\textbf{Reference}
&
\textbf{Metric}
&
\textbf{Covering radius}
&
\textbf{Entropy bound}
&
\textbf{$n$-free}
\\
\midrule

\text{Alon et al.}
\cite{Alon1997-scale-sensitive}
&
$\ell_\infty$
&
$2\gamma$
&
\[
\log N_\infty(V,2\gamma)
=
O\left(
d
\log\frac{en}{\gamma}
\log\frac{en}{d\gamma}
\right)
\]
&
No
\\

Bartlett, Kulkarni, and Posner
\cite{BartlettKulkarniPosner1997}
&
$\ell_1$
&
$C\gamma$
&
\[
\log N_1(V,C\gamma)
=
O\left(
d\log^2\frac{e}{\gamma}
\right)
\]
&
Yes
\\

Bartlett and Long
\cite{BARTLETT1998}
&
$\ell_1$
&
$\gamma+\varepsilon$
&
\[
\log N_1(V,\gamma+\varepsilon)
=
O\left(
d\log\frac{1}{\varepsilon}
\log^2(en)
\right)
\]
&
No
\\

Mendelson and Vershynin
\cite{MendelsonVershynin2003}
&
$\ell_2$
&
$C\gamma$
&
\[
\log N_2(V,C\gamma)
=
O\left(
d\log\frac{e}{\gamma}
\right)
\]
&
Yes
\\

Rudelson and Vershynin
\cite{rudelson2006combinatorics}
&
$\ell_\infty$
&
$C\gamma/\rho$
&
\[
\log N_\infty
\left(
V,\frac{C\gamma}{\rho}
\right)
=
O\left(
d
\log\frac{en}{d\gamma}
\log^\rho\frac{en}{d}
\right)
\]
&
No
\\

\text{Aiyer et al.}
\cite{aiyer2026scale}
&
$\ell_\infty$
&
$\gamma/2+\varepsilon$
&
\[
\log N_\infty
\left(
V,\frac{\gamma}{2}+\varepsilon
\right)
=
O\left(
d\log\frac{1}{\varepsilon}\log^2(en)
\right)
\]
&
No
\\

This work
&
$\ell_2$
&
$\gamma/2+\varepsilon$
&
\[
\log N_2
\left(
V,\frac{\gamma}{2}+\varepsilon
\right)
=
O\left(
d\log^3\frac{e}{\varepsilon}
\right)
\]
&
Yes
\\

\bottomrule
\end{tabularx}
\caption{
Comparison of covering-number bounds for bounded real-valued
function classes. Throughout, $d=\fat_\gamma(V)$,
$\varepsilon,\rho\in(0,1)$, and $C>0$ denotes a universal
constant. The family $V\subseteq[-1,+1]^n$ is fixed, and $N_p(V,r)$ denotes its covering number in the indicated normalized metric.
}
\label{tab:covering-bounds}
\end{table}

\begin{technicalresult}
\begin{theorem}\label{Thm:ell_2-bound}
Let $n\geq1$, let $V\subseteq[-1,+1]^n$, let $\gamma>0$, and write $d=\fat_\gamma(V)<\infty$. For every $\varepsilon\in(0,1)$,
\[
\log N_2\left(V,\frac{\gamma}{2}+\varepsilon\right)
\leq C d\log^3\frac{e}{\varepsilon},
\]
where $C>0$ is a universal constant.
\end{theorem}
\end{technicalresult}
The proof constructs real-valued covers by iteratively constructing covers for partial binary classes using Theorem~\ref{Thm:Par-Cover}.
We overview this argument in more detail in 
\Cref{Sec:Overview-Covers}. The full \hyperref[Proof:Cover-ell_2]{proof} appears in Section~\ref{Sec:Technical-Proofs}.

It remains open whether our factor $\log^3(1/\varepsilon)$ can be reduced to $\log(1/\varepsilon)$ while retaining radius $\gamma/2+\varepsilon$. A single logarithm suffices at radius $C\gamma$, by the classical result of Mendelson and Vershynin \cite{MendelsonVershynin2003}, which settled Talagrand's entropy problem.

The bound depends on $n$ only through $d=\fat_\gamma(V)$. This is the covering bound that we combine with the chaining argument in the next subsection to prove our uniform-convergence result.

\subsection{Chaining at optimal scale}\label{Sec:Refined-Chaining}

We now turn to the last part of our technical contribution: a chaining bound at an optimal scale. Chaining is a standard and powerful technique used to transform covering bounds into uniform convergence bounds. We first introduce Rademacher complexity, the intermediate quantity in this approach, then recall classical chaining and explain our contribution.

\paragraph{Rademacher complexity.}
Consider a family $V\subseteq[-1,+1]^n$ of vectors. Color each coordinate independently red or blue, with equal probability. For any fixed vector, the red and blue sums are equal in expectation. The vector with the largest red-minus-blue discrepancy, however, may depend on the coloring. Rademacher complexity is the expected maximum of this signed discrepancy, normalized by $n$. Writing $\sigma_i=+1$ for red and $\sigma_i=-1$ for blue, define

\[
\Rad(V)
=
\Ex_\sigma\left[
\sup_{v\in V}\frac1n\sum_{i=1}^n\sigma_i v_i
\right].
\]
For a finite family, the following lemma bounds its Rademacher complexity using the lengths of the vectors and the size of~$V$.

\begin{lemma}[Massart's lemma, see Lemma 26.8 in \cite{understandingml}]\label{Lem:Massart}
Let $V\subseteq\R^n$ be finite and nonempty, and suppose that every $v\in V$ has normalized $\ell_2$ norm at most $R$.\footnote{Recall that the normalized Euclidean norm is $\|v\|_2=(n^{-1}\sum_{i=1}^n v_i^2)^{1/2}$.} Then
\[
\Rad(V)\leq R\sqrt{\frac{2\log|V|}{n}}.
\]
\end{lemma}

\paragraph{From finite covers to chaining.}
What about infinite families $V$?
A finite cover of radius $r$ approximates every vector in $V$ with error at
most $r$. We can therefore bound $\Rad(V)$ by applying Massart's lemma
to the cover and adding $r$ for the approximation error. Chaining
refines this argument: it uses covers at successively smaller radii
and applies Massart's lemma to the differences between successive
approximations.
Let $H_i$ be covers of $V$ at decreasing radii $r_i$, starting with $H_0=\{0\}$ and $r_0=1$. For each $v\in V$, choose $v_i\in H_i$ with $\|v-v_i\|_2\leq r_i$. We write
\[
v=(v-v_k)+\sum_{i=1}^k(v_i-v_{i-1}).
\]
At each level, the possible differences $v_i-v_{i-1}$ form a finite family, with at most $|H_i||H_{i-1}|$ members. Massart's lemma bounds their random imbalance using their lengths and the logarithm of their number. Summing these bounds over the levels, and adding the residual error, gives a chaining bound. A classical form is the following.

\begin{theorem}[Dudley's chaining bound \cite{NIPS2010_76cf99d3}]
For every $n\geq1$, every $V\subseteq[-1,+1]^n$, and every $\alpha\in(0,1)$,
\[
\Rad(V)
\leq
4\alpha+
\frac{10}{\sqrt n}\int_0^1
\sqrt{\log N_2(V,\alpha+t)}\,dt.
\]
\end{theorem}

The cutoff $\alpha$ allows us to use covers only at radii above $\alpha$, at the cost of the leading term $4\alpha$.
The coefficient in front of the leading term cannot be smaller than $1$. Indeed, consider the family $V=\{-\alpha,+\alpha\}^n$. The zero function covers this class at radius $\alpha$, so the entropy integral vanishes. Yet $\Rad(V)=\alpha$: for each coloring $\sigma$, the class contains the vector $\alpha\sigma$. This leaves the question: \emph{can a chaining bound attain the leading term $\alpha$, using only covers at radii above $\alpha$?}

This is critical in our application. Here, the covering bounds hold at every radius above $\gamma/2$, so we take $\alpha=\gamma/2$. The symmetrization step that converts Rademacher complexity into uniform convergence introduces another factor of two. The classical bound would therefore give a leading error of $4\gamma$, whereas we need~$\gamma$. Increasing the sample size does not remove this loss: we need to reduce the coefficient of $\alpha$ from $4$ to~$1$.

\paragraph{Aligning the covers.}
To see the difficulty, recall the successive representatives $v_i$ chosen from covers of radii $r_i$. When these covers are chosen independently, the triangle inequality gives
\[
\|v_i-v_{i-1}\|_2\leq r_i+r_{i-1}.
\]
As the radii decrease to $\alpha>0$, this estimate approaches $2\alpha$. Thus, increasingly accurate approximations need not give increasingly small bounds on the successive differences. This is where controlling only the sizes of the covers falls short.

We instead choose the covers jointly, allowing them to be larger in order to align their representatives. Here, alignment means that each vector has a sequence of representatives that retain the prescribed approximation radii, while their successive differences become small as the radii approach~$\alpha$. These smaller differences let us apply Massart's lemma without paying an additional constant multiple of $\alpha$. We explain the construction, and how we control the increase in cover size, in the proof overview in Section~\ref{Sec:Overview-Chaining}.

\begin{technicalresult}
\begin{theorem}[Chaining at optimal scale]\label{Thm:Chain-Imp}
For every $n\geq1$, every $V\subseteq[-1,+1]^n$, and every $\alpha\in(0,1)$,
\[
\Rad(V)
\leq
\alpha+\frac{32}{\sqrt n}
\int_0^1\sqrt{\log N_2(V,\alpha+t^2)}\,dt.
\]
\end{theorem}
\end{technicalresult}

The leading term has the optimal coefficient $1$. The entropy integral uses radii $\alpha+t^2$ rather than $\alpha+t$; our covering bounds keep it finite. The complete \hyperref[Proof:Chain-Imp]{proof} appears in Section~\ref{Sec:Chaining-Proof}.

\section{Proof Overview}\label{Sec:Proof-Over}

This section gives an overview of the proof of our uniform-convergence bound, focusing on the main ideas behind the argument. We begin with the lower-bound construction and then turn to the upper bound, where the covering and chaining tools are combined. For the geometric margin-net results in Theorems~\ref{Thm:Net-Hilbert-Halfspaces} and~\ref{Thm:Net-Hamming-Ball}, we refer the reader directly to their complete proofs in Section~\ref{Sec:Margin-Nets-Proofs}.


For a class $\F\subseteq[-1,+1]^\X$ and a sample $S=(x_1,\ldots,x_n)$ from a distribution $\D$, write
\[
\Rep_\D(\F,S)
=\sup_{f\in\F}
\left|\Ex_{x\sim\D}f(x)-\frac1n\sum_{i=1}^n f(x_i)\right|.
\]
Thus, $m_\gamma(\varepsilon,\delta)$ is the least sample size $m$ such that, for every $n\geq m$ and every $\D$, this error is at most $\gamma+\varepsilon$ with probability at least $1-\delta$. We set it to infinity if no such $m$ exists. The class satisfies $\gamma$-uniform convergence when this sample complexity is finite for every $\varepsilon>0$ and $\delta\in(0,1)$.

\subsection{Lower bound}\label{Sec:Overview-Lower}

The construction separates the two sources of error. We start with a binary class $\H$ of VC dimension $d$ for which uniform convergence requires $\Omega((d+\log(1/\delta))/\varepsilon^2)$ samples: with fewer samples, some distribution gives an estimation error greater than $\varepsilon$ with probability greater than $\delta$. We choose $\H$ to be symmetric, meaning that $-h\in\H$ whenever $h\in\H$. Add a coordinate $t\in[0,1]$ and consider the functions
\[
f_{h,b}(t,x)
=\left(1-\frac\gamma2\right)h(x)+\frac\gamma2 b(t),
\qquad h\in\H,
\]
where $b$ ranges over all (measurable) functions from $[0,1]$ to $(-1,{+1})$.

The added term can vary by strictly less than $\gamma$, so it cannot itself realize the two labels in a fat-shattering pattern at scale $\gamma$. The resulting class still has fat-shattering dimension $d$ at this scale. Nevertheless, this term contributes an error of $\gamma$: under the uniform distribution on $[0,1]$, a function~$b$ can be arbitrarily close to $-1$ on all sampled points and arbitrarily close to $+1$ everywhere else.

Taking the product with a hard distribution for $\H$ makes these two effects independent. Since $h$ and $b$ can be chosen separately, their worst-case errors add.
Achieving error $\gamma+\varepsilon$ therefore requires estimating the binary class to accuracy of order $\varepsilon$. The classical binary lower bound supplies the required dependence on $d$, $\varepsilon$, and $\delta$. The complete construction and its analysis appear in the \hyperref[prop:lower-bnd]{proof of Proposition~\ref*{prop:lower-bnd}} in Section~\ref{Sec:Main-Result-Proof}.

\subsection{Upper bound}\label{Sec:Overview-Upper}

We first show how the covering and chaining bounds from Section~\ref{Sec:Technical-Contributions} imply the upper bound. We then explain the construction of the covers in Section~\ref{Sec:Overview-Covers} and their alignment for chaining in Section~\ref{Sec:Overview-Chaining}.

Let $\F\subseteq[-1,+1]^\X$ and write $d=\fat_\gamma(\F)$. For a sample $S=(x_1,\ldots,x_n)$, consider the family of vectors
\[
V_S=\{(f(x_1),\ldots,f(x_n)):f\in\F\}\subseteq[-1,+1]^n.
\]
Thus, $\fat_\gamma(V_S)\leq d$ and the results of Section~\ref{Sec:Technical-Contributions} apply to every sample $S$. The covering bound in Theorem~\ref{Thm:ell_2-bound} gives
\[
\log N_2(V_S,\gamma/2+\tau)
\leq C d\log^3(e/\tau),
\]
and chaining at optimal scale (Theorem~\ref{Thm:Chain-Imp}), with $\alpha=\gamma/2$, yields
\[
\begin{aligned}
\Rad(V_S)
&\leq\frac\gamma2+
\frac{32}{\sqrt n}
\int_0^1\sqrt{\log N_2(V_S,\gamma/2+t^2)}\,dt\\
&\leq\frac\gamma2+
C'\sqrt{\frac dn}\int_0^1\log^{3/2}(e/t)\,dt
=\frac\gamma2+O\!\left(\sqrt{\frac dn}\right).
\end{aligned}
\]
Since this bound holds for every sample, classical symmetrization \cite[Lemma~26.2]{understandingml} and McDiarmid's concentration inequality \cite{mcdiarmid1989method} give, with probability at least $1-\delta$,
\[
\Rep_\D(\F,S)
\leq\gamma+
O\!\left(\sqrt{\frac{d+\log(2/\delta)}n}\right).
\]
Taking $n$ of order $(d+\log(2/\delta))/\varepsilon^2$ makes this error at most $\gamma+\varepsilon$. The formal argument appears in the \hyperref[Thm:sharp-bound]{proof of Theorem~\ref*{Thm:sharp-bound}} in Section~\ref{Sec:Main-Result-Proof}. We now explain the two ingredients in this calculation: constructing covers arbitrarily close to radius $\gamma/2$, and chaining these covers without enlarging the leading error.

\subsubsection{Constructing covers}\label{Sec:Overview-Covers}

\paragraph{Partial binary classes.}
\begingroup
\emergencystretch=1em
We construct approximate disambiguations using the fractional-cover rounding approach from Section~\ref{Sec:Margin-Nets}. Let
\par\endgroup
\[
\F\subseteq\{-1,{+1},\star\}^{\X}
\]
be a partial concept class of VC dimension \(d\). We first construct a distribution over total binary functions under which every \(f\in\F\) has a nonnegligible probability of being approximated, and then round this fractional cover into a finite one.

A natural starting point is the learnability of \(\F\). Given \(f\in\F\), sample sufficiently many points from \(\X\) uniformly and apply a learner to the resulting \(f\)-labeled sample. With constant probability, the output approximates \(f\) on all but an \(\varepsilon\)-fraction of \(\X\). To obtain a distribution that does not depend on \(f\), one might instead sample the  points, assign each of them an independent uniformly random label from \(\{-1,{+1},\star\}\), and apply the learner to the resulting labeled sample. For any fixed \(f\), however, the probability of guessing all  labels correctly is \(3^{-m}\), where $m$ is the number of sampled points. This is too small for the subsequent rounding argument.

The loss comes from guessing the entire labeled sample, although the learner can be reconstructed from much less information. We therefore construct a compression scheme of size
\[
O\left(d\log\frac{e}{\varepsilon}\right).
\]
Rather than guessing all \(m\) labels, we guess only a compression record and the labels of the examples it contains, and then apply the reconstruction map. Controlling the effective number of possible compression records shows that the resulting distribution assigns probability at least
\[
\left(\frac{c}{\varepsilon}\right)^{-C d\log(e/\varepsilon)}
\]
to functions that \(\varepsilon\)-approximate any fixed \(f\in\F\). This gives the required fractional disambiguation.

%
\begingroup
Finally, we round the fractional cover. For each $f\in\F$, let $B_\alpha(f)$ be the set of total binary functions that disagree with $f$ on at most an $\alpha$-fraction of $\X$, counting only coordinates where $f$ is defined. The $\star$ entries impose no constraints, so these sets are not ordinary Hamming balls and Theorem~\ref{Thm:Net-Hamming-Ball} does not apply directly. Nevertheless, the same rounding strategy applies, using the $\varepsilon$-net theorem for partial concept classes of Bourneuf, Charbit, and Thomass{\'e}~\cite{Dense2025Bourneuf}.

Increasing the allowed error from $\alpha$ to $\alpha+\eta$ supplies the margin: the associated partial class distinguishes error at most $\alpha$ from error at least $\alpha+\eta$, leaving intermediate errors undefined. This class has VC dimension $O(1/\eta^2)$, so the $\eps$-net theorem for partial concept classes  gives a finite cover with error at most $\alpha+\eta$. Taking $\alpha=\eta=\varepsilon/2$ and substituting the fractional-cover bound yields the desired result.
\par\endgroup

\paragraph{Covering real-valued classes.}

We now overview the construction of the covers in Theorem~\ref{Thm:ell_2-bound}.
For $V\subseteq[-1,+1]^n$ with $d=\fat_\gamma(V)$, we seek a cover
at radius $\gamma/2+\varepsilon$ whose size is independent of~$n$.
We first recall the standard approach of obtaining covers from packing
bounds \cite{haussler1995sphere,dudley1978central}, and explain why it
does not preserve the optimal scale $\gamma/2$. Start with a packing
of $M$ vectors whose pairwise normalized $\ell_2$ distances are at
least $\rho$. The idea resembles the Johnson--Lindenstrauss lemma:
reduce the dimension while preserving this packing, with a small
loss in separation. Here, the reduction simply
restricts the vectors to randomly sampled coordinates. For two vectors
at normalized $\ell_2$ distance at least $\rho$, sampling
$O(1/\rho^2)$ coordinates preserves a constant fraction of their
distance with constant probability. Increasing the sample to
$O(\rho^{-2}\log M)$ allows concentration and a union bound to
preserve the separation of all pairs simultaneously.

We thus retain all $M$ vectors in a packing on only logarithmically
many coordinates. Provided the retained separation is sufficiently
large relative to $\gamma$, known bounds in terms of the fat-shattering
dimension limit this projected packing to $M^{o(1)}$ vectors. Since the projected packing still contains $M$ vectors, we obtain
$M\leq M^{o(1)}$. This inequality fails for sufficiently large $M$,
giving an upper bound on $M$ independent of $n$.

The difficulty is preserving the radius. A maximal packing by
radius-$r$ balls gives a cover at radius~$2r$. This route therefore
incurs a factor-two loss, whereas we need a radius arbitrarily close
to $\gamma/2$. We instead construct the covers directly, adapting
the iterative approach of Aiyer et al.~\cite{aiyer2026scale}.

We now describe one iteration of the cover construction.
Suppose the current cover approximates every $v\in V$ within
$\gamma/2+s$ on all but a small fraction of coordinates.
We replace each representative~$g$ by a small collection that
reduces the excess error from $s$ to $s/2$.
Different vectors represented by $g$ may require different corrections;
our partial-class bound supplies a small list of correction patterns
that serves them all.
In more detail, for each vector $v$ approximated by $g$, we record which coordinates
of $g$ should be increased or decreased:
\[
p_v(i)=
\begin{cases}
+1,&v_i-g_i\geq\gamma/2,\\
-1,&v_i-g_i\leq-\gamma/2,\\
\star,&\text{otherwise}.
\end{cases}
\]
The undefined coordinates already have error less than $\gamma/2$.
For fixed $g$, these partial vectors form a class of VC dimension
at most $d$: shattering coordinates would fat-shatter them in $V$. Theorem~\ref{Thm:Par-Cover} therefore gives
a small approximate disambiguation $H_g$. Replace $g$ by
\[
g+\frac{s}{2}h,\qquad h\in H_g,
\]
clipping the resulting representatives to $[-1,+1]$.

For each $v$, choose an $h$ that approximately disambiguates $p_v$.
Where the previous approximation was accurate and $h$ respects the
defined label, the correction reduces the error to at most
$\gamma/2+s/2$. On undefined coordinates, either direction satisfies
this bound. Thus the new representatives halve the excess error,
apart from the coordinates where the previous approximation or the
disambiguation was inaccurate.

Starting from $\{0\}$, we repeat this step $O(\log(1/\varepsilon))$
times. We choose the disambiguation errors small enough that the
total fraction of inaccurate coordinates remains small. Since all
values lie in $[-1,+1]$, these coordinates contribute little to
normalized $\ell_2$ distance, giving radius $\gamma/2+\varepsilon$.
Each round adds $O(d\log^2(1/\varepsilon))$ to the logarithm of the
cover size; the number of rounds supplies the third logarithmic factor:
\[
\log N_2(V,\gamma/2+\varepsilon)
=O\!\left(d\log^3(1/\varepsilon)\right).
\]

\subsubsection{Chaining at optimal scale}\label{Sec:Overview-Chaining}

We now turn to the proof of the chaining theorem
(Theorem~\ref{Thm:Chain-Imp}). Recall that our goal is to bound the
Rademacher complexity using covers at radii approaching \(\alpha\) from
above, while retaining the optimal leading term \(\alpha\).
We follow the chaining argument recalled in
Section~\ref{Sec:Refined-Chaining}: write each vector as its final
approximation error plus the successive differences between its
representatives, and apply Massart's lemma to those differences.
Our modification concerns the choice of representatives. We align
them so that successive differences become small even when the
approximation radii approach a positive limit $\alpha$.
To see why this is needed, consider covers $H_i$ of
$V\subseteq[-1,+1]^n$ at radii $r_i$ decreasing to $\alpha$.
If $h_i(v)$ and $h_{i-1}(v)$ are successive representatives of
the same vector $v$, the triangle inequality gives
\[
\|h_i(v)-h_{i-1}(v)\|_2\leq r_i+r_{i-1}.
\]
When the radii tend to $\alpha$, this bound tends to
$2\alpha$. The following geometric observation allows us to align
the representatives while preserving their approximation radii.
Suppose that \(y\) is the representative already chosen for \(v\) at radius \(R\), and \(x\) is a representative from the next cover, whose radius is \(r<R\). Then
\[
v\in B_R(y)\cap B_r(x).
\]
The next representative only needs to cover the vectors in this intersection; it need not cover the entire ball \(B_r(x)\). This gives us room to move its center toward \(y\) while keeping the same radius \(r\).

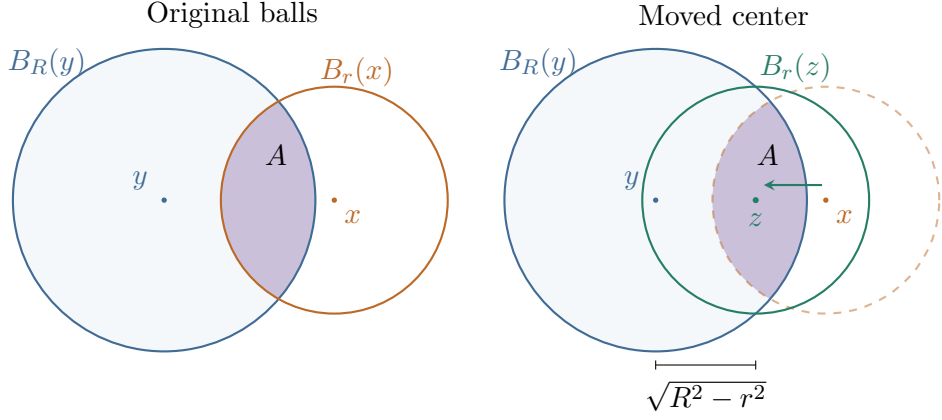
\begin{figure}
    \centering
    \begin{tikzpicture}
\definecolor{chainBlue}{RGB}{64,107,148}
\definecolor{chainOrange}{RGB}{187,105,38}
\definecolor{chainGreen}{RGB}{38,125,102}
\definecolor{chainPurple}{RGB}{159,141,187}
\foreach \offset in {0,6.5} {
    \begin{scope}[xshift=\offset cm]
      \fill[chainBlue!5] (0,0) circle (2);
      \begin{scope}
        \clip (0,0) circle (2);
        \fill[chainPurple!55] (2.25,0) circle (1.5);
      \end{scope}
      \draw[chainBlue,thick] (0,0) circle (2);
      \fill[chainBlue] (0,0) circle (.035);
      \node[chainBlue,above left] at (-.08,0) {$y$};
      \node[chainBlue] at (-1.55,1.82) {$B_R(y)$};
      \fill[chainOrange] (2.25,0) circle (.035);
      \node[chainOrange,below right] at (2.25,0) {$x$};
      \node at (1.48,.6) {$A$};
    \end{scope}
  }
  \node at (.9,2.45) {Original balls};
  \draw[chainOrange,thick] (2.25,0) circle (1.5);
  \node[chainOrange] at (2.55,1.7) {$B_r(x)$};
  \begin{scope}[xshift=6.5cm]
    \node at (.9,2.45) {Moved center};
    \draw[chainOrange,dashed,opacity=.5,thick] (2.25,0) circle (1.5);
    \draw[chainGreen,thick] (1.322876,0) circle (1.5);
    \fill[chainGreen] (1.322876,0) circle (.04);
    \node[chainGreen,below] at (1.322876,-.04) {$z$};
    \node[chainGreen] at (1.85,1.75) {$B_r(z)$};
    \draw[->,>=stealth,chainGreen,thick] (2.2,.2) -- (1.43,.2);
    \draw (0,-2.18) -- (1.322876,-2.18);
    \draw (0,-2.12) -- (0,-2.24);
    \draw (1.322876,-2.12) -- (1.322876,-2.24);
    \node[below] at (.661438,-2.25) {$\sqrt{R^2-r^2}$};
  \end{scope}
\end{tikzpicture}
    \caption{Moving the center while preserving the approximation. The shaded intersection \(A=B_R(y)\cap B_r(x)\) remains covered when we replace \(B_r(x)\) by \(B_r(z)\). The radius stays \(r\), while the new center satisfies \(\|z-y\|_2\leq\sqrt{R^2-r^2}\).}
    \label{Fig:Chaining-Geometry}
\end{figure}

The following Euclidean fact makes the improvement quantitative (Lemma~\ref{Lem:Geo-center}): the intersection \(B_R(y)\cap B_r(x)\) is contained in a ball \(B_r(z)\) whose center satisfies
\[
\|z-y\|_2
\leq
\sqrt{R^2-r^2}.
\]
Unlike \(R+r\), this quantity tends to zero when the two radii approach the same positive value; see Figure~\ref{Fig:Chaining-Geometry}.

Applying this operation successively produces representatives \(\pi_i(v)\) that remain within distance \(r_i\) of \(v\), while satisfying
\[
\|\pi_i(v)-\pi_{i-1}(v)\|_2
\leq
\sqrt{r_{i-1}^2-r_i^2}.
\]
Thus, although the approximation radii converge only to \(\alpha\), the successive increments converge to zero.

Alignment increases the cover sizes because a center may move to
different positions depending on the preceding choices. Choosing
\(r_i=\alpha+4^{-i}\)
makes the successive differences shrink geometrically, with norms
$O(2^{-i})$; the resulting increase in the Massart bounds changes
only the universal constant in the entropy term, while the leading
approximation error remains $\alpha$.
Stopping after $N$ scales leaves approximation error at most $r_N$.
Applying Massart's lemma to the successive differences and summing
their contributions gives
\[
\Rad(V)
\leq
r_N
+
\frac{C}{\sqrt n}
\sum_{i=1}^N
2^{-i}\sqrt{\log |H_i|},
\]
where $H_i$ is the original cover at radius $r_i$,
before alignment.
Choosing each $H_i$ to have minimum size, bounding the sum by an
integral, and letting $N$ tend to infinity gives the desired
entropy bound.

\section{Proofs}\label{Sec:Technical-Proofs}
This section contains the complete proofs of Our results. In section \ref{Sec:Margin-Nets-Proofs} we prove the $\varepsilon$-nets results such as Theorems \ref{Thm:Net-Hamming-Ball}, and \ref{Thm:Net-Hilbert-Halfspaces}. In section \ref{Sec:Chaining-Proof} We prove theorem the chaining bound of Theorem \ref{Thm:Chain-Imp}. In section \ref{Sec:Partial-Covers-Proof} We prove results on partial concept classes to derive the covering bound of Theorem \ref{Thm:Par-Cover}. We apply this result in section \ref{Sec:Real-Covers-Proof} to derive the covering bounds of Theorem \ref{Thm:ell_2-bound}. Finally in section \ref{Sec:Main-Result-Proof} we use the chaining and covering results to prove our main theorem.

Throughout this section, the constants $C,C',c>0$ are universal and may change from line to line.  

\subsection{\texorpdfstring{$\varepsilon$}{ε}-Nets and Fractional Covers with Margin Proofs}\label{Sec:Margin-Nets-Proofs}

The following is Theorem 18 of  \cite{Dense2025Bourneuf}, we repeat the proof here for completeness.
\begin{lemma}[$\delta$-net for partial concept classes {\cite{Dense2025Bourneuf}}]\label{Lem:Nets-Par}
    Let $\F\subseteq \{-1,{+1},\star\}^{\X}$ be a partial concept class with
    $\VC(\F)=d<\infty$. Let $\delta\in(0,1)$, and let $\D$ be a distribution
    over $\X$ such that, for every $f\in\F$,
    \[
    \Pr_{x\sim\D}\bigl(f(x)=1\bigr)\geq \delta.
    \]
    Then there exists a set $X\subseteq\X$ of size
    \[
    |X|\leq C\frac{d}{\delta}\log\frac{e}{\delta}+1.
    \]
    such that, for every $f\in\F$, there is some $x\in X$ for which
    $f(x)\neq -1$; that is, $f(x)\in\{1,\star\}$.
\end{lemma}

\begin{proof}
The proof is similar to the one in Theorem 18 of \cite{Dense2025Bourneuf}, and is given here for completeness. If $d=0$, the result is clear so we may assume $d>0$.
   Assume toward contradiction no such set exists.
    Let $n$ be chosen later, draw $x_1,\ldots,x_n$ independently from $\D$,
    and independently assign each $x_i$ a uniformly random label
    $y_i\in\{-1,{+1}\}$. Let
    \[
    S=\{(x_i,y_i)\}_{i=1}^n,
    \]
    and let $A$ be the event that $S$ is realizable by $\F$.

    For every fixed choice of $x_1,\ldots,x_n$, Sauer's lemma implies that
    $\F$ realizes at most
    \[
    \sum_{j=0}^d\binom{n}{j}
    \leq \left(\frac{en}{d}\right)^d
    \]
    labelings of these points. Since the labels are uniform,
    \[
    \Pr(A)\leq 2^{-n}\left(\frac{en}{d}\right)^d.
    \]

    We now generate the same random sample in a different order. First choose
    a random subset $I\subseteq[n]$ by including each index independently
    with probability $1/2$. The points indexed by $I$ receive label $-1$,
    while the remaining points receive label $1$. Write
    \[
    X=\{x_i:i\in I\},
    \qquad K=|I|.
    \]
    By assumption there exists some $f\in\F$ such that
    \[
    f(x)=-1
    \qquad\text{for every }x\in X.
    \]
    Fix such $I,f$, and note that if $f(x_i)=1$ for every $i\neq I$, then $S$ is realizable by $f$ so $A$ occurs. For each such point $f(x_i)=1$ with probability at least $\delta$, hence
    \[
    \Pr(A\mid X)\geq \delta^{\,n-K}.
    \]
    Taking expectations  
    \[
    \Pr(A)
    \geq \Ex\!\left[\delta^{\,n-K}\right]=\frac{1}{2^n} \sum_{k=0}^n\binom{n}{k}\delta^{n-k}=\frac{(1+\delta)^n}{2^n}.
    \]
   So we have \[
   (1+\delta)^n\leq 2^n\Pr(A)\leq \left(\frac{en}{d}\right)^d.
   \]
   Solving for $n$ in the above, we get that for \[
   n=C\frac{d}{\delta}\log \frac{e}{\delta},
   \]
   the above  will not hold, from which we deduce the desired result.
\end{proof}

\begin{rtheorem}{Thm:Net-Hamming-Ball}\label{Proof:Net-Hamming-Ball}
Let $\mu$ be a probability distribution on $\{0,1\}^n$. Fix a radius $\alpha$ and an additive margin $\eta$ satisfying $0<\eta\leq\alpha$ and $\alpha+\eta\leq1$. For every $\varepsilon\in(0,1)$, there is a set $S\subseteq\{0,1\}^n$ of size
\[
    |S|=O\left(\frac{\alpha^2}{\varepsilon\eta^2}\log\frac{e}{\varepsilon}\right)
\]
{such that every closed Hamming ball of radius $\alpha$ and $\mu$-mass at least $\varepsilon$ intersects $S$ after its radius is increased to $\alpha+\eta$.}

The bound is optimal up to the logarithmic factor for small radii. For every sufficiently small $\varepsilon,\alpha>0$ and every $0<\eta\leq\alpha$, there are a dimension $n$ and a distribution $\mu$ for which every such net has $\Omega(\alpha^2/(\varepsilon\eta^2))$ points.
\end{rtheorem}

\begin{proof}[Proof of Theorem \ref{Thm:Net-Hamming-Ball}]

Set $\X=\{0,1\}^n$, and consider the partial concept class
$\mathcal{H}=
\{
h_x\; :\; x\in \X\}\subseteq\{-1,{+1},\star\}^{\X}$ given by
\[
    h_x(y)
    =
    \begin{cases}
        +1, & d_H(x,y)\leq\alpha,\\
        \star, & \alpha<d_H(x,y)\leq\alpha+\eta,\\
        -1, & d_H(x,y)>\alpha+\eta,
    \end{cases}
    \qquad x,y\in\X.
\]
We first show that
\[
    \VC(\mathcal{H})
    =
    O\left(\frac{\alpha^2}{\eta^2}\right).
\]
It is enough to bound shattered sets whose cardinality is a power of
two, since every set of size $m\geq2$ contains a subset of
power-of-two size greater than $m/2$. Suppose, therefore, that
$u_1,\ldots,u_d$ are shattered by $\H$, where $d$ is a power of two,
and let $A$ be a Hadamard matrix of order $d$.

Choose $x,y\in\{-1,+1\}^d$ maximizing $\inP{x}{Ay}$, and set
\[
a_{i,j}=A_{i,j}x_i y_j.
\]
By shattering, for each $j\in[d]$ there is some
$v_j\in\{-1,+1\}^n$ such that
\[
d_H(u_i,v_j)
\begin{cases}
\leq \alpha, & a_{i,j}=+1,\\
\geq \alpha+\eta, & a_{i,j}=-1.
\end{cases}
\]
For each $i\in[d]$, define $u_i'$ by setting $u_i'(k)$ to be $1$ if $u_i(k)=1$, and $-1$ otherwise. Define $v_j'$ in the same way.
Since
\[
d_H(u_i,v_j)
=
\frac12-\frac{1}{2n}\inP{u_i'}{v_j'},
\]
we have
\[
2\sum_{i,j}a_{i,j}d_H(u_i,v_j)
=
\frac1n\sum_{k=1}^n
\left(
\sum_{i,j}a_{i,j}
-
\sum_{i,j}a_{i,j}u_i'(k)v_j'(k)
\right)
\geq 0.
\]
Indeed, for each $k$, replacing $x_i$ by $x_i u_i'(k)$ and $y_j$ by
$y_jv_j'(k)$ produces another pair of sign vectors, so the inequality
follows from the maximality of $x$ and $y$.

On the other hand, since $A$ is a Hadamard matrix,
\[
\sum_{i,j}a_{i,j}
=
\inP{x}{Ay}
\leq
\lVert x\rVert_2\lVert Ay\rVert_2
=
d\sqrt d.
\]
Let $P$ and $N$ denote the numbers of pairs $(i,j)$ for which
$a_{i,j}=+1$ and $a_{i,j}=-1$, respectively. Then
\[
P+N=d^2,
\qquad
P-N\leq d\sqrt d,
\]
while the choice of the $v_j$ gives
\[
0
\leq
\sum_{i,j}a_{i,j}d_H(u_i,v_j)
\leq
P\alpha-N(\alpha+\eta).
\]
Consequently,
\[
1+\frac{\eta}{\alpha}
\leq
\frac{P}{N}
\leq
\frac{d^2+d\sqrt d}{d^2-d\sqrt d}
=
1+\frac{2}{\sqrt d-1}.
\]
It follows that
\[
d
\leq
\left(1+\frac{2\alpha}{\eta}\right)^2
\leq
9\frac{\alpha^2}{\eta^2},
\]
where the last inequality uses $\eta\leq\alpha$. Since every shattered
set contains a power-of-two subset of at least half its size, we
conclude that
\[
\VC(\H)
=
O\left(\frac{\alpha^2}{\eta^2}\right).
\]

Restrict $\mathcal{H}$ to those $h_x$ for which
\[
    \mu\bigl(h_x^{-1}(1)\bigr)
    =
    \mu\bigl(B_\alpha(x)\bigr)
    \geq\varepsilon.
\]
Lemma \ref{Lem:Nets-Par} now gives a set
$S\subseteq\X$ of size
\[
    |S|
    =
    O\left(
        \frac{\alpha^2}{\eta^2\varepsilon}
        \log\frac{e}{\varepsilon}
    \right)
\]
that intersects
\[
    h_x^{-1}(\{1,\star\})
    =
    B_{\alpha+\eta}(x)
\]
for every such $x$. This proves the upper bound.

%
%

For the lower bound, we use the matrix form of Spencer's discrepancy theorem \cite{spencer1985six}: there is a universal constant $C_0$ such that, for any collection of at most $m$ vectors in $[-1,+1]^m$, some $z\in\{-1,+1\}^m$ has inner product of absolute value at most $2C_0\sqrt m$ with every vector in the collection.

Let $\psi_t:\{0,1\}^m\to\{0,1\}^{tm}$ repeat each coordinate $t$ times. The discrepancy bound implies that, for any set $T\subseteq\{0,1\}^{tm}$ with $|T|\leq m$, there is an $x\in\{0,1\}^m$ such that
\[
    D_H(\psi_t(x),s)\geq t\left(\frac m2-C_0\sqrt m\right)
    \qquad\text{for every }s\in T,
\]
where $D_H$ denotes unnormalized Hamming distance. Indeed, for each $s\in T$, average its signs $2s_j-1$ within each block of $t$ coordinates to obtain a vector in $[-1,+1]^m$, and apply Spencer's theorem to these vectors. Thus repeating coordinates preserves the lower bound even when the net points do not themselves have repeated coordinates.

We prove the claim for every $0<\alpha\leq1/32$, every $0<\eta\leq\alpha$, and all sufficiently small $\varepsilon>0$. Write $q=\alpha/\eta$. First suppose $q\geq q_0$, where $q_0$ is a sufficiently large universal constant, and set
\[
    m=\lfloor q^2/16\rfloor,
    \qquad
    r=\left\lfloor\frac m2-(C_0+2)\sqrt m\right\rfloor.
\]
We choose $q_0$ large enough that $r\geq m/4$. By the central limit theorem, there is a universal constant $p_0>0$ such that
\[
    2^{-m}\sum_{j=0}^{r}\binom mj\geq p_0.
\]
Let $k=\lfloor p_0/\varepsilon\rfloor$. We place $k$ copies of the repeated cube on the same active coordinates and separate them using additional code coordinates. Repetition lets us provide enough code coordinates while keeping $\alpha$ and $\eta$ at their prescribed values.

Choose a sufficiently large positive integer $t$ and put
\[
    N=\left\lceil\frac{tr}{\alpha}\right\rceil,
    \qquad L=N-tm.
\]
Since $r\geq m/4$ and $\alpha\leq1/32$, we have $L\geq7N/8$. By taking $t$ large enough, a standard random-coding argument gives $k$ words $c_1,\ldots,c_k\in\{0,1\}^L$ whose pairwise distances are at least $L/4$. Define
\[
    \mathcal F_i=\{(\psi_t(x),c_i):x\in\{0,1\}^m\},
    \qquad
    \mathcal F=\bigcup_{i=1}^k\mathcal F_i,
\]
and let $\mu$ be uniform on $\mathcal F$. For every $u\in\mathcal F_i$, the ball of normalized radius $\alpha$ around $u$ contains all points in that copy whose underlying $m$-bit words differ from $u$ in at most $r$ coordinates, since $tr\leq\alpha N$. Hence
\[
    \mu(B_\alpha(u))\geq\frac{p_0}{k}\geq\varepsilon.
\]

In unnormalized distance, the enlarged radius is $R=(\alpha+\eta)N$. The rounding in the definition of $N$ gives
\[
    \frac Rt
    <\left(1+\frac1q\right)\left(r+\frac\alpha t\right)
    <\frac m2-C_0\sqrt m,
\]
where the last inequality uses $\sqrt m/q\leq1/4$ and $\alpha/t\leq1/32$. Also, $R\leq N/16$, so distinct copies have distance at least $L/4>2R$.

Let $S$ be a net with the required property, and let $S_i$ consist of its points at distance at most $R$ from $\mathcal F_i$. The sets $S_i$ are pairwise disjoint, and each must cover $\mathcal F_i$ by radius-$R$ balls. If $|S_i|\leq m$, applying the repeated-coordinate discrepancy bound to the first $tm$ coordinates of these points produces a point of $\mathcal F_i$ farther than $R$ from all of them, a contradiction. Therefore
\[
    |S|\geq km
    =\Omega\left(\frac{q^2}{\varepsilon}\right)
    =\Omega\left(\frac{\alpha^2}{\varepsilon\eta^2}\right).
\]

Finally, if $1\leq q<q_0$, take $k=\lfloor1/\varepsilon\rfloor$ points in a sufficiently large binary cube with pairwise normalized distances at least $1/4$, and put the uniform distribution on them. Each radius-$\alpha$ ball centered at one of these points has mass at least $1/k\geq\varepsilon$. Their enlargements are disjoint because $2(\alpha+\eta)\leq4\alpha\leq1/8$. Every net therefore needs at least $k=\Omega(1/\varepsilon)=\Omega(q^2/\varepsilon)$ points, since $q_0$ is a universal constant.

We conclude by explaining the extension to $\ell_p$. The only property
of the Hamming metric used in the upper bound is
\[
\sum_{i,j}a_{i,j}d_H(u_i,v_j)\geq 0.
\]
The analogous inequality holds for arbitrary points in $L_1$. Indeed,
every $L_1$ metric is an integral of cut metrics:
\[
\begin{aligned}
\sum_{i,j}a_{i,j}\lVert u_i-v_j\rVert_1
&=
\int_{\Omega}\int_{\mathbb R}
\sum_{i,j}a_{i,j}
\left|
\mathbf{1}_{\{u_i(s)\geq t\}}
-
\mathbf{1}_{\{v_j(s)\geq t\}}
\right|
\,dt\,d\mu(s)\\
&\geq 0.
\end{aligned}
\]
For each $s$ and $t$, the inner sum is nonnegative by the same
maximality argument used above for binary vectors. The remainder of the
proof is therefore unchanged. For $1<p\leq2$,  we use \cite{Bretagnolle1966} to embed  $\ell_p$ isometrically, up to a common
scaling factor, into $L_1$. Since this scaling affects the radius and
margin equally, the same upper bound follows for every
$p\in[1,2]$.

The lower-bound construction can likewise be viewed inside $\ell_p$
using the normalized embedding of the Hamming cube, under which
\[
\lVert x-y\rVert_p=d_H(x,y)^{1/p}.
\]
An $\ell_p$ radius $\alpha$ and additive margin $\eta$ then correspond
to Hamming radius $\alpha^p$ and Hamming margin
$(\alpha+\eta)^p-\alpha^p$. Since
\[
\frac{\alpha^p}{(\alpha+\eta)^p-\alpha^p}
\asymp
\frac{\alpha}{\eta}
\qquad
\text{for }0<\eta\leq\alpha\text{ and }1\leq p\leq2,
\]
the construction gives the same lower bound
\[
\Omega\left(\frac{\alpha^2}{\varepsilon\eta^2}\right)
\]
for strong $\varepsilon$-nets in $\ell_p$. This argument does not
extend to weak nets, whose points may lie outside the embedded Hamming
cube.
\end{proof}

\begin{rtheorem}{Thm:Net-Hilbert-Halfspaces}
\label{Proof:Net-Hilbert-Halfspaces}
Let $A$ be a set of $n$ points in the unit ball of a real Hilbert space $E$, and let $\varepsilon,\eta\in(0,1)$. There is a set $S\subseteq E$ of size $O(1/(\varepsilon\eta^2))$ such that every closed halfspace containing at least $\varepsilon n$ points of $A$ has distance at most $\eta$ from some point of $S$.

The bound is optimal: for all sufficiently small $\varepsilon,\eta>0$, there is a finite set $A$ in the unit ball of $\ell_2$ for which every such net has $\Omega(1/(\varepsilon\eta^2))$ points.

\end{rtheorem}

\begin{proof}[Proof of Theorem~\ref{Thm:Net-Hilbert-Halfspaces}]
For the upper bound, let $\mu$ be the uniform distribution on $A$, and let $X\sim\mu$. For a unit vector $u\in E$ and $b\in\R$, write
\[
    H(u,b)=\{x\in E:\inP{u}{x}\geq b\}.
\]
Consider the second-moment operator $T=\mathbb{E}[X\otimes X]$. This operator has finite rank, since $X$ takes only finitely many values, so the spectral argument below applies even when $E$ is infinite-dimensional. Since $A$ lies in the unit ball,
\[
    \operatorname{tr}(T)=\mathbb{E}\|X\|^2\leq1.
\]
Let $V$ be the span of the eigenvectors of $T$ whose eigenvalues are larger than $\varepsilon\eta^2/2$, and let $P$ denote the orthogonal projection onto $V$. Then
\[
    r:=\dim(V)\leq\frac{2}{\varepsilon\eta^2}.
\]
For every unit vector $u\in E$,
\[
    \mathbb{E}\left[\inP{P^\perp u}{X}^2\right]
    =\inP{TP^\perp u}{P^\perp u}
    \leq\frac{\varepsilon\eta^2}{2}.
\]
Consequently,
\[
    \Pr\left(\inP{P^\perp u}{X}>\eta\right)\leq\frac{\varepsilon}{2}.
\]
If $\mu(H(u,b))\geq\varepsilon$, there is therefore some $x\in A$ with
\[
    \inP{u}{x}\geq b
    \qquad\text{and}\qquad
    \inP{P^\perp u}{x}\leq\eta.
\]
It follows that
\[
    \inP{u}{Px}=\inP{u}{x}-\inP{P^\perp u}{x}\geq b-\eta.
\]
Since $\|Px\|\leq1$, the enlarged halfspace $H(u,b-\eta)$ intersects the unit ball of $V$.

Choose a simplex $\Delta=\operatorname{conv}\{v_0,\ldots,v_r\}\subseteq V$ containing the unit ball of $V$, and set $S=\{v_0,\ldots,v_r\}$. A linear functional attains its maximum over a simplex at a vertex. Thus, whenever $H(u,b-\eta)$ intersects the unit ball of $V$, some $v_i$ satisfies $\inP{u}{v_i}\geq b-\eta$. As $u$ is a unit vector, the distance from $v_i$ to $H(u,b)$ is at most $\eta$. Hence the closed ball of radius $\eta$ centered at $v_i$ intersects $H(u,b)$, as required. Finally,
\[
    |S|=r+1\leq\frac{2}{\varepsilon\eta^2}+1.
\]
The same argument applies to any probability distribution $\mu$ supported on the unit ball of a finite-dimensional Euclidean space, with the condition $|A\cap H|\geq\varepsilon n$ replaced by $\mu(H)\geq\varepsilon$.

For the lower bound, set $p=1/16$ and fix $\varepsilon\in(0,p/2)$ and $\eta\in(0,1/32)$. Let
\[
    m=\left\lfloor\frac{1}{256\eta^2}\right\rfloor,
    \qquad
    k=\left\lfloor\frac{p}{\varepsilon}\right\rfloor.
\]
Then $m=\Theta(\eta^{-2})$ and $\eta\leq1/(16\sqrt m)$.
Let $E=E_1\oplus\cdots\oplus E_k$ be an orthogonal direct sum of $m$-dimensional spaces. Choose an orthonormal basis $e_{i1},\ldots,e_{im}$ of each $E_i$, and take the scaled cube vertices
\[
    A_i=\left\{\frac{1}{\sqrt m}\sum_{j=1}^m\sigma_j e_{ij}:\sigma\in\{-1,+1\}^m\right\}.
\]
The set $A=\bigcup_{i=1}^k A_i$ consists of $n=k2^m$ points on the unit sphere. Let $\mu_i$ be uniform on $A_i$, so that $\mu=k^{-1}\sum_i\mu_i$ is uniform on $A$.

We first show that, for every unit vector $z\in E_i$,
\[
    \mu_i\left(\left\{x:\inP{z}{x}\geq\frac{1}{4\sqrt m}\right\}\right)\geq p.
\]
Indeed, if $X_i\sim\mu_i$ and $Z=\sqrt m\,\inP{z}{X_i}$, then $Z$ is symmetric and
\[
    \mathbb{E}Z^2=1,
    \qquad
    \mathbb{E}Z^4=3-2\sum_{j=1}^m\inP{z}{e_{ij}}^4\leq3.
\]
Cauchy--Schwarz gives
\[
    1=\mathbb{E}Z^2
    \leq\frac{1}{16}+\sqrt{3\Pr(|Z|\geq1/4)}.
\]
By symmetry,
\[
    \Pr(Z\geq1/4)=\frac12\Pr(|Z|\geq1/4)
    \geq\frac{(1-1/16)^2}{6}>p.
\]

Now let $S\subseteq E$ satisfy the required intersection property, and put $W=\operatorname{span}(S)$. Write $P_W$ for the orthogonal projection onto $W$. We claim that, for every $i\in[k]$,
\[
    \sum_{j=1}^m\|P_W e_{ij}\|^2\geq\frac m2.
\]
Otherwise, some unit basis vector $e\in E_i$ satisfies $\|P_W e\|^2<1/2$. Define
\[
    v=\frac{P_{W^\perp}e}{\|P_{W^\perp}e\|}.
\]
Then $v$ is a unit vector orthogonal to $W$, and
\[
    \|P_{E_i}v\|\geq\inP{v}{e}
    =\|P_{W^\perp}e\|>\frac{1}{\sqrt2}.
\]
Apply the preceding estimate to $z=P_{E_i}v/\|P_{E_i}v\|$. For $x\in A_i$ with $\inP{z}{x}\geq1/(4\sqrt m)$, we have
\[
    \inP{v}{x}>\frac{1}{4\sqrt{2m}}>2\eta.
\]
Thus $\mu_i(H(v,2\eta))\geq p$ and
\[
    \mu(H(v,2\eta))\geq\frac pk\geq\varepsilon.
\]
Yet every $s\in S$ satisfies $\inP{v}{s}=0$, so its distance from $H(v,2\eta)$ is $2\eta$. No closed ball of radius $\eta$ centered at a point of $S$ intersects this halfspace, a contradiction.

Finally, the vectors $e_{ij}$ form an orthonormal basis of $E$, and hence
\[
    |S|\geq\dim(W)=\operatorname{tr}(P_W)
    =\sum_{i=1}^k\sum_{j=1}^m\|P_W e_{ij}\|^2
    \geq\frac{km}{2}
    =\Omega\left(\frac{1}{\varepsilon\eta^2}\right).
\]
Finally, embed $E$ isometrically in $\ell_2$. Any set of centers in $\ell_2$ satisfying the theorem's conclusion can be projected onto $E$ without changing its inner products with vectors in $E$ or increasing its cardinality. The same lower bound therefore holds in $\ell_2$ for all sufficiently small $\varepsilon,\eta>0$.
\end{proof}



\subsection{Chaining at optimal scale}\label{Sec:Chaining-Proof}
We work with a fixed family $V\subseteq[-1,+1]^n$ and the normalized Euclidean norm. We use $\Rad(V)$ and $N_2(V,r)$ as defined in Sections~\ref{Sec:Refined-Chaining} and~\ref{Sec:Real-Covers}; the same definition of Rademacher complexity applies to the finite families in $\R^n$ that arise below.
We also recall Massart’s lemma
\begin{rlemma}{Lem:Massart}
Let $V\subseteq\R^n$ be finite and nonempty, and suppose that every $v\in V$ has normalized $\ell_2$ norm at most $R$. Then
\[
\Rad(V)\leq R\sqrt{\frac{2\log|V|}{n}}.
\]
\end{rlemma}

\begin{lemma}\label{Lem:Geo-center}
    Let $R>r>0$ be two positive numbers, and let $B_r(x)$, $B_R(y)$ be two balls in a Euclidean space. Set 
    \[
    A=B_r(x)\cap B_R(y).
    \]
    Then there exists $z$ such that $d(y,z)\leq \sqrt{ R^2-r^2}$, and $A\subseteq B_r(z)$. 
    
\end{lemma}
\begin{proof}
By translation, assume that $y=0$. If $\norm{x}\leq\sqrt{R^2-r^2}$, take $z=x$. Otherwise, set
\[
 z=\frac{\sqrt{R^2-r^2}}{\norm{x}}x.
\]
Fix $a\in A$. If $\inP{z}{a-z}\leq0$, then since $x-z=\lambda z$ for some $\lambda>0$, we have $\inP{x-z}{z-a}\geq0$. Consequently,
\[
 \norm{x-a}^2
 =\norm{x-z}^2+\norm{z-a}^2+2\inP{x-z}{z-a}
 \geq\norm{a-z}^2.
\]
Since $a\in B_r(x)$, this gives $a\in B_r(z)$. If instead $\inP{z}{a-z}>0$, then
\[
 \norm{a}^2
 =\norm{z}^2+\norm{a-z}^2+2\inP{z}{a-z}
 \geq\norm{z}^2+\norm{a-z}^2.
\]
Using $\norm{a}\leq R$ and $\norm{z}^2=R^2-r^2$, we again obtain $\norm{a-z}^2\leq r^2$.

\end{proof}

\begin{lemma}[Refined Chaining]\label{Lem:Chain-Cover}
Let $V\subseteq[-1,+1]^n$ and $\alpha\in(0,1)$. For every integer $N\geq1$, there is a finite $(\alpha+4^{-N})$-cover $\G\subseteq\R^n$ of $V$ satisfying
\[
\Rad(\G)\leq\frac{32}{\sqrt n}\int_0^1
\sqrt{\log N_2(V,\alpha+t^2)}\,dt.
\]
\end{lemma}
\begin{proof}
Set $\alpha_i=\alpha+4^{-i}$ for $0\leq i\leq N$, and let $H_0=\{0\}$. For $1\leq i\leq N$, choose an $\alpha_i$-cover $H_i$ of $V$ of minimum size. For each $v\in V$, fix $h_i(v)\in H_i$ within distance $\alpha_i$ of $v$.

We construct approximations $\pi_i(v)$ satisfying
\[
\|v-\pi_i(v)\|_2\leq\alpha_i,
\qquad
\|\pi_i(v)-\pi_{i-1}(v)\|_2
\leq\sqrt{\alpha_{i-1}^2-\alpha_i^2}.
\]
Start with $\pi_0(v)=0$. Given $\pi_i(v)$, apply Lemma~\ref{Lem:Geo-center} to
\[
A=B_{\alpha_i}(\pi_i(v))\cap B_{\alpha_{i+1}}(h_{i+1}(v)).
\]
It gives a center $\pi_{i+1}(v)$ such that
\[
A\subseteq B_{\alpha_{i+1}}(\pi_{i+1}(v)),
\qquad
\|\pi_{i+1}(v)-\pi_i(v)\|_2
\leq\sqrt{\alpha_i^2-\alpha_{i+1}^2}.
\]
Since $v\in A$, both required inequalities follow. Fix the choice of the new center for each pair $(\pi_i(v),h_{i+1}(v))$, so it depends only on this pair. Consequently,
\[
|\pi_i(V)|\leq\prod_{j=1}^i|H_j|.
\]
Set $\G=\pi_N(V)$. This is an $(\alpha+4^{-N})$-cover of $V$.

To bound its Rademacher complexity, write each representative as a sum of successive differences,
\[
\pi_N(v)=\sum_{i=1}^N\bigl(\pi_i(v)-\pi_{i-1}(v)\bigr),
\]
and let $\Delta_i=\{\pi_i(v)-\pi_{i-1}(v):v\in V\}$. Subadditivity of the supremum gives
\[
\Rad(\G)\leq\sum_{i=1}^N\Rad(\Delta_i).
\]
Every vector in $\Delta_i$ has norm at most $\sqrt{\alpha_{i-1}^2-\alpha_i^2}$, and
\[
|\Delta_i|\leq|\pi_i(V)|\,|\pi_{i-1}(V)|
\leq\prod_{j=1}^i|H_j|^2.
\]
Thus Massart's lemma and subadditivity of the square root yield
\[
\Rad(\Delta_i)
\leq\frac{2\sqrt{\alpha_{i-1}^2-\alpha_i^2}}{\sqrt n}
\sum_{j=1}^i\sqrt{\log|H_j|}.
\]
Since $\alpha<1$,
\[
\sqrt{\alpha_{i-1}^2-\alpha_i^2}
=\sqrt{(4^{1-i}-4^{-i})(2\alpha+4^{1-i}+4^{-i})}
\leq4\cdot2^{-i}.
\]
Summing in the opposite order and using $\sum_{i=j}^N2^{-i}\leq2\cdot2^{-j}$ gives
\[
\begin{aligned}
\Rad(\G)
&\leq\frac8{\sqrt n}\sum_{j=1}^N\sum_{i=j}^N2^{-i}\sqrt{\log|H_j|}\\
&\leq\frac{32}{\sqrt n}\sum_{j=1}^N2^{-j-1}
\sqrt{\log N_2(V,\alpha+4^{-j})}.
\end{aligned}
\]
By monotonicity of the covering number, each summand is at most the integral of $\sqrt{\log N_2(V,\alpha+t^2)}$ over $[2^{-j-1},2^{-j}]$. These intervals are disjoint subsets of $[0,1]$, proving the claim.
\end{proof}

\begin{rtheorem}{Thm:Chain-Imp}\label{Proof:Chain-Imp}
For every $n\geq1$, every $V\subseteq[-1,+1]^n$, and every $\alpha\in(0,1)$,
\[
\Rad(V)\leq\alpha+
\frac{32}{\sqrt n}\int_0^1
\sqrt{\log N_2(V,\alpha+t^2)}\,dt.
\]
\end{rtheorem}
\begin{proof}[Proof of Theorem~\ref{Thm:Chain-Imp}]
For every integer $N\geq1$, Lemma~\ref{Lem:Chain-Cover} gives an $(\alpha+4^{-N})$-cover $\G$ of $V$ with the stated bound on $\Rad(\G)$. By Cauchy--Schwarz, replacing a vector by a representative within normalized $\ell_2$ distance $\alpha+4^{-N}$ changes its normalized signed sum by at most $\alpha+4^{-N}$. Hence
\[
\Rad(V)\leq\alpha+4^{-N}+\Rad(\G)
\leq\alpha+4^{-N}+
\frac{32}{\sqrt n}\int_0^1
\sqrt{\log N_2(V,\alpha+t^2)}\,dt.
\]
Letting $N\to\infty$ proves the theorem.
\end{proof}

\subsection{Approximate Disambiguation via Compression}\label{Sec:Partial-Covers-Proof}

\begin{definition}[Loss for partial labels]
For $y\in\{-1,{+1},\star\}$ and a total function
$h\in\{-1,{+1}\}^{\X}$, define
\[
\ell_{x,y}(h)
=
\begin{cases}
0, & y=\star,\\
\mathbf{1}_{\{h(x)\neq y\}}, & y\in\{-1,{+1}\}.
\end{cases}
\]
Thus, for a partial function
$f\in\{-1,{+1},\star\}^{\X}$ and a finite set $U\subseteq\X$,
\[
L_{U,f}(h)
=
\frac{1}{|U|}
\sum_{x\in U}\ell_{x,f(x)}(h)
=
\frac{1}{|U|}
\bigl|
\{x\in U\cap\operatorname{supp}(f):h(x)\neq f(x)\}
\bigr|.
\]
Similarly, if
\[
S=((x_i,y_i))_{i=1}^m
\]
is a labeled sample and $\D$ is a distribution over
$\X\times\{-1,{+1},\star\}$, set
\[
L_S(h)
=
\frac{1}{m}\sum_{i=1}^m\ell_{x_i,y_i}(h),
\qquad
L_{\D}(h)
=
\Ex_{(x,y)\sim\D}\ell_{x,y}(h).
\]
\end{definition}

\begin{definition}[$\varepsilon$-compression scheme]
Let
\[
\F\subseteq\{-1,{+1},\star\}^{\X}.
\]
An $\varepsilon$-compression scheme of size $k$ for $\F$
consists of a compressor $\kappa$ and a reconstructor $\rho$
with the following properties.

For every sample
\[
S=((x_i,y_i))_{i=1}^n,
\]
the compressor returns a sequence
\[
\kappa(S)=I=(i_1,\ldots,i_k)\in[n]^k.
\]
Writing
\[
S_I=((x_{i_j},y_{i_j}))_{j=1}^k,
\]
the reconstructor outputs a total function
\[
\rho(S_I)\in\{-1,{+1}\}^{\X}.
\]
Moreover, whenever $S$ is realizable by some $f\in\F$, we have
\[
L_S\bigl(\rho(S_I)\bigr)\leq\varepsilon.
\]

For a given $n$, we say that the scheme has effective size
$s_n$ if there exists a set
\[
\mathcal{I}_n\subseteq[n]^k,
\qquad
|\mathcal{I}_n|\leq s_n,
\]
such that, for every sample $S$ of size $n$, there is some
$J\in\mathcal{I}_n$ satisfying
\[
\rho(S_{\kappa(S)})=\rho(S_J).
\]
\end{definition}

\begin{theorem}[$\varepsilon$-compression for VC classes]
\label{Thm:Compress-VC}
Let $\F\subset{\{-1,+1,\star\}}^{\X}$ be a partial concept class of VC dimension
$d=\VC(\F){\geq1}$.\footnote{If $d=0$, a fixed total completion of the class gives a scheme with $k=0$ and $s_n=1$.} Then, for every $\varepsilon\in(0,1)$, $\F$ admits an
$\varepsilon$-compression scheme of size
\[
k
\leq
C d\log\frac{e}{\varepsilon},
\]
where $C>0$ is a universal constant. Moreover, for every $n>d$, the
compression scheme has effective size
\[
s_n
\leq
\left(\frac{en}{d}\right)^{
C d\log\frac{e}{\varepsilon}
}.
\]

\end{theorem}

\begin{proof}
Fix a finite sample
\[
S=\bigl((x_i,f(x_i))\bigr)_{i=1}^n
\]
realizable by some $f\in\F$. We use the partial concept class learner
of \cite{alon2022theory}. There is a total-valued learner $A$ such that,
for every distribution $\D$ supported on $S$, there is a sequence $B$
of at most $Cd$ examples from $S$ for which
\[
L_{\D}(A(B))\leq\frac{1}{3}.
\]
The learner may be taken to be invariant under permutations and
repetitions of its input\footnote{This was not formally proved in \cite{alon2022theory} but is clear from the definition of the learner as a one-inclusion-graph based learner.}.

Strictly speaking, the learner in \cite{alon2022theory} is stated for
binary-labeled samples. This causes no difficulty here: before applying
the learner, we simply remove the $\star$-labeled examples. Since these
examples contribute zero loss for every total hypothesis, the same
weak-learning guarantee holds with respect to our loss.

By the minimax theorem, there is a distribution $\mu$ over such
sequences $B$ such that, for every distribution $\D$ supported on $S$,
\[
\Ex_{B\sim\mu}L_{\D}(A(B))
\leq
\frac{1}{3}.
\]
In particular, by taking $\D$ to be a point mass, we obtain, for every
$x_i$ such that $f(x_i)\neq\star$,
\[
\Pr_{B\sim\mu}\bigl(A(B)(x_i)\neq f(x_i)\bigr)
\leq
\frac{1}{3}.
\]

Let
\[
T
=
\left\lceil
C\log\frac{e}{\varepsilon}
\right\rceil,
\]
draw independently
\[
B_1,\ldots,B_T\sim\mu,
\]
and define
\[
h
=
\Maj\bigl(A(B_1),\ldots,A(B_T)\bigr).
\]
For every $x_i$ such that $f(x_i)\neq\star$, Hoeffding's inequality
gives
\[
\begin{aligned}
\Pr\bigl(h(x_i)\neq f(x_i)\bigr)
&\leq
\Pr\left[
\frac{1}{T}
\sum_{t=1}^T
\mathbf{1}_{\{A(B_t)(x_i)\neq f(x_i)\}}
\geq
\frac{1}{2}
\right] \\
&\leq
e^{-cT}
\leq
\varepsilon,
\end{aligned}
\]
where $c>0$ is a universal constant. Since $\star$-labeled examples
contribute zero loss, it follows that
\[
\Ex_{B_1,\ldots,B_T}L_S(h)
\leq
\varepsilon.
\]
Consequently, there is a deterministic choice of
$B_1,\ldots,B_T$ satisfying
\[
L_S\left(
\Maj\bigl(A(B_1),\ldots,A(B_T)\bigr)
\right)
\leq
\varepsilon.
\]

The compressor chooses one such tuple and records the concatenation of
the corresponding index sequences. By padding each block with
repetitions if necessary, we may assume that all blocks have the same
fixed length. The reconstructor splits the compressed sequence into
these blocks, applies $A$ to each block, and returns their majority
vote. The total number of recorded examples is at most
\[
k
\leq
C dT
\leq
C d\log\frac{e}{\varepsilon}.
\]

It remains to bound the effective size. Since $A$ is invariant under
permutations and repetitions, the hypothesis $A(B_t)$ depends only on
the set of distinct sample indices appearing in $B_t$. For $n>d$, the
number of possible outputs of a single block is therefore at most
\[
\left(\frac{en}{d}\right)^{Cd},
\]
after adjusting the universal constant. Since there are
\[
T
\leq
C\log\frac{e}{\varepsilon}
\]
blocks, the number of possible reconstructed hypotheses is at most
\[
s_n
\leq
\left(\frac{en}{d}\right)^{CdT}
\leq
\left(\frac{en}{d}\right)^{
C d\log\frac{e}{\varepsilon}
}.
\]
\end{proof}

\begin{theorem}[Sample complexity for compression-based learners]
\label{Thm:compres-sample-complex}
Let
\[
\F\subseteq\{-1,{+1},\star\}^{\X}
\]
admit an $\alpha$-compression scheme of size $k$ and effective
size $s_n$. Let $\D$ be realizable by $\F$, draw
\[
S\sim\D^n,
\]
where $n>2k$, and set
\[
m=n-k.
\]
Then, with probability at least $1-\delta$,
\[
L_{\D}\bigl(\rho(S_{\kappa(S)})\bigr)
\leq
2\alpha
+
\sqrt{
\frac{4\alpha\log(s_n/\delta)}{m}
}
+
\frac{4\log(s_n/\delta)}{m}.
\]
\end{theorem}

\begin{proof}
Let
\[
\mathcal{I}_n=\{I_1,\ldots,I_{s_n}\}
\]
be the collection in the definition of effective size, and set
\[
h_j=\rho(S_{I_j}).
\]
Let $J_j\subseteq[n]$ be the set of distinct indices appearing
in $I_j$, and let
\[
V_j=[n]\setminus J_j.
\]
Since $I_j$ has length $k$,
\[
|V_j|\geq n-k=m.
\]

For each fixed $j$, the function $h_j$ depends only on the
examples indexed by $J_j$. Conditional on these examples, the
remaining sample
\[
S_{V_j}
\]
consists of $|V_j|$ independent samples from $\D$ and is
independent of $h_j$. Therefore, Bernstein's inequality implies
that, with probability at least $1-\delta/s_n$,
\[
L_{\D}(h_j)
\leq
L_{S_{V_j}}(h_j)
+
\sqrt{
\frac{
2L_{S_{V_j}}(h_j)\log(s_n/\delta)
}{
|V_j|
}
}
+
\frac{
4\log(s_n/\delta)
}{
|V_j|
}.
\]
A union bound shows that this inequality holds simultaneously
for every $j\leq s_n$ with probability at least $1-\delta$.

Write
\[
h=\rho(S_{\kappa(S)}).
\]
By the definition of effective size, there is some $j\leq s_n$
such that
\[
h=h_j.
\]
Since the compression scheme has empirical error at most
$\alpha$ on $S$,
\[
L_S(h_j)\leq\alpha.
\]
It follows that
\[
L_{S_{V_j}}(h_j)
\leq
\frac{n}{|V_j|}L_S(h_j)
\leq
\frac{n}{m}\alpha
\leq
2\alpha,
\]
where the last inequality uses $n>2k$. Moreover,
\[
|V_j|\geq m.
\]
Substitution into the Bernstein bound therefore gives
\[
L_{\D}(h)
\leq
2\alpha
+
\sqrt{
\frac{4\alpha\log(s_n/\delta)}{m}
}
+
\frac{4\log(s_n/\delta)}{m}.
\]
\end{proof}

\begin{corollary}[Sample complexity for VC classes via compression]
\label{Cor:Compres-lrn}
Let
\[
\F\subseteq\{-1,{+1},\star\}^{\X}
\]
be a partial concept class of VC dimension $d\geq 1$. There is
a compression-based learner such that, for every
$\varepsilon,\delta\in(0,1)$, if
\[
n
\geq
\frac{C}{\varepsilon}
\left(
d\log^2\frac{e}{\varepsilon}
+
\log\frac{1}{\delta}
\right),
\]
then, for every distribution $\D$ realizable by $\F$, with
probability at least $1-\delta$ over $S\sim\D^n$,
\[
L_{\D}\bigl(\rho(S_{\kappa(S)})\bigr)
\leq
\varepsilon.
\]
In particular, for constant confidence, the sample complexity
is
\[
O\left(
\frac{d}{\varepsilon}
\log^2\frac{e}{\varepsilon}
\right).
\]
\end{corollary}

\begin{proof}
Apply Theorem~\ref{Thm:Compress-VC} with compression error
\[
\alpha=\frac{\varepsilon}{8}.
\]
This gives a compression scheme of size
\[
k
\leq
C d\log\frac{e}{\varepsilon}.
\]
For $n>2k$, its effective size satisfies
\[
\log s_n
\leq
C d\log\frac{e}{\varepsilon}
\log\frac{en}{d}.
\]
Consequently,
\[
\log\frac{s_n}{\delta}
\leq
C d\log\frac{e}{\varepsilon}
\log\frac{en}{d}
+
\log\frac{1}{\delta}.
\]

We claim that the assumed lower bound on $n$, with a sufficiently large universal constant, implies
\[
 \frac{\log(s_n/\delta)}{n}\leq c\varepsilon
\]
for a sufficiently small universal constant $c>0$. Write $L=\log(e/\varepsilon)$ and $b=\log(1/\delta)$, and let
\[
 n_0=\left\lceil\frac{K}{\varepsilon}(dL^2+b)\right\rceil.
\]
For $n\geq n_0\geq d$, the function
\[
 \frac{dL\log(en/d)+b}{n}
\]
is decreasing, so it suffices to bound it at $n_0$. The inequality $\log(1+t)\leq t$ gives
\[
 \log(en_0/d)\leq O(L+\log K)+\frac{b}{dL^2}.
\]
It follows that $\log(s_n/\delta)/n\leq O((\log K)/K)\varepsilon$. Choosing $K$ sufficiently large proves the claim and also ensures $n>2k$, hence $m=n-k\geq n/2$.


Theorem~\ref{Thm:compres-sample-complex} now gives
\[
\begin{aligned}
L_{\D}\bigl(\rho(S_{\kappa(S)})\bigr)
&\leq
2\alpha
+
\sqrt{
\frac{4\alpha\log(s_n/\delta)}{m}
}
+
\frac{4\log(s_n/\delta)}{m} \\
&\leq
\frac{\varepsilon}{4}
+
\sqrt{C_1c}\,\varepsilon
+
C_2c\varepsilon.
\end{aligned}
\]
Choosing $c$ sufficiently small makes the right-hand side at
most $\varepsilon$, which proves the result.
\end{proof}

\begin{lemma}[Fractional covering for partial function classes]
\label{Lem:Par-Frac-cover}
Let $\F\subset \{-1,{+1},\star\}^{\X}$ be a partial concept class over a
finite domain $\X$, and assume that $\VC(\F)=d<\infty$. Then, for every
$\varepsilon\in(0,1)$, there exists a distribution $\mu$ over
$\{-1,{+1}\}^{\X}$ such that, for every $f\in\F$,
\[
\Pr_{h\sim\mu}\bigl(L_{\X,f}(h)\leq\varepsilon\bigr)
\geq
\left(\frac{c}{\varepsilon}\right)^{
-Cd\log\frac{e}{\varepsilon}
},
\]
where $C>0$ and $c>1$ are universal constants.
\end{lemma}

\begin{proof}
The case $d=0$ is trivial, so assume that $d\geq 1$. By
Theorem~\ref{Thm:Compress-VC} and
Corollary~\ref{Cor:Compres-lrn}, after adjusting the universal
constants, there is an $\varepsilon$-compression scheme
$(\kappa,\rho)$ for $\F$ of size
\[
k
\leq
Cd\log\frac{e}{\varepsilon}
\]
such that, for
\[
m
=
\left\lceil
\frac{Cd}{\varepsilon}
\log^2\frac{e}{\varepsilon}
\right\rceil,
\]
the corresponding compression-based learner has loss at most
$\varepsilon$ with probability at least $1/2$ when applied to an
$m$-sample from the uniform distribution on $\X$. Moreover, the
effective size of the compression scheme satisfies
\[
s_m
\leq
\left(\frac{em}{d}\right)^{
Cd\log\frac{e}{\varepsilon}
}.
\]
Thus, there is a collection
\[
\mathcal I_m\subseteq[m]^k,
\qquad
|\mathcal I_m|\leq s_m,
\]
such that, for every labeled sample $S$ of size $m$, there is some
$I\in\mathcal I_m$ for which
\[
\rho(S_I)=\rho(S_{\kappa(S)}).
\]

We now define the distribution $\mu$. First draw
\[
x_1,\ldots,x_m
\]
independently and uniformly from $\X$. Next, draw
\[
I=(i_1,\ldots,i_k)
\]
uniformly from $\mathcal I_m$. Finally, draw
\[
y_1,\ldots,y_k
\]
independently and uniformly from $\{-1,{+1},\star\}$, and output
\[
h
=
\rho\bigl((x_{i_j},y_j)_{j=1}^k\bigr).
\]
The law of the resulting total function $h$ defines $\mu$ and does
not depend on the choice of $f\in\F$.

Fix $f\in\F$, and let
\[
S
=
\bigl((x_i,f(x_i))\bigr)_{i=1}^m.
\]
By the choice of $m$, with probability at least $1/2$ over the draw of
$x_1,\ldots,x_m$,
\[
L_{\X,f}\bigl(\rho(S_{\kappa(S)})\bigr)
\leq
\varepsilon.
\]
Condition on such a choice of $x_1,\ldots,x_m$. By the definition of
effective size, there is some $I\in\mathcal I_m$ such that
\[
\rho(S_I)=\rho(S_{\kappa(S)}).
\]
The randomly chosen index list equals one such $I$ with probability
at least $1/s_m$. Conditional on this event, the random labels satisfy
\[
y_j=f(x_{i_j})
\qquad
\text{for every }j\in[k]
\]
with probability $3^{-k}$. Therefore,
\[
\Pr\left[
h=\rho(S_{\kappa(S)})
\,\middle|\,
x_1,\ldots,x_m
\right]
\geq
\frac{1}{3^ks_m}.
\]
It follows that
\[
\Pr_{h\sim\mu}\bigl(L_{\X,f}(h)\leq\varepsilon\bigr)
\geq
\frac{1}{2\cdot 3^ks_m}.
\]

Using the bounds on $k$ and $s_m$, we obtain
\[
3^ks_m
\leq
\left(\frac{3em}{d}\right)^{
Cd\log\frac{e}{\varepsilon}
}.
\]
Moreover,
\[
\frac{m}{d}
\leq
\frac{C}{\varepsilon}
\log^2\frac{e}{\varepsilon}\leq \frac{c}{\varepsilon^3}
\]

Consequently, after adjusting the universal constants
\[
\frac{1}{2\cdot 3^ks_m}
\geq
\left(\frac{c}{\varepsilon}\right)^{
-Cd\log\frac{e}{\varepsilon}
}.
\]
This proves the result.
\end{proof}

\begin{lemma}[From Fractional Cover to Finite Cover]\label{Lem:Par-Frac-to-Cover}
{Let $\F\subseteq\{-1,+1,\star\}^{\X}$ be a partial function class on a finite domain $\X$, and let $\alpha,\delta\in(0,1)$. Assume there is a distribution $\D$ over $\{-1,+1\}^{\X}$ such that, for each $f\in\F$,}
   \[
   \Pr_{h\sim \D}[L_{\X,f}(h)\leq \alpha]\geq \delta.
   \]
   Then, for any $ {\varepsilon\in(0,1)}$ there is some $H\subset {\{-1,+1\}}^\X$, of size
   \[
   |H|\leq \frac{C}{\varepsilon^2 \delta}\log \frac{e}{\delta},
   \]
 Such that for any $f\in \F$, there is some $h\in H$ with 
   \[
   L_{\X,f}(h)\leq \alpha+\varepsilon.
   \]
\end{lemma}

\begin{proof}
Let $\H$ be the support of $\D$, and for each $f\in \F$ define the partial function $w_f:\H\to \{-1,{+1},\star\}$ by
 \[
w_f(h)=\begin{cases}
    1 \quad& L_{\X,f}(h)\leq \alpha,
    \\-1 \quad& L_{\X,f}(h)\geq \alpha+\varepsilon,
    \\\star \quad& \text{otherwise}.
\end{cases}
\]
Set $\W=\{w_f\;:\; f\in \F\}$.
Let $f\in \F$, $h\in \H$. Treat the $\star$ as $0$, 
and note that under this identification, $|f(x)|-f(x)h(x)$, is $2$ if $f(x)$ is defined (nonzero) and not equal $h(x)$, and it is $0$ if either $f(x)=h(x)$, or $f(x)=0$. Summing over this we get 
\[
{L_{\X,f}(h)=\frac{1}{2|\X|}\sum_{x\in\X}\bigl(|f(x)|-f(x)h(x)\bigr)=\frac{|\supp(f)|-\inP{f}{h}}{2|\X|},}
\]
where $\supp(f)$ is the set of points for which $f(x)\neq0$.
Hence the class $\W=\{w_f \;:\; f\in \F\}$ can be embedded as the class of affine classifiers with margin. Formally define \[
\varphi_f=\frac{-f}{2\sqrt{|\X|}}, \quad s_f= \frac{|\supp(f)|}{2|\X|}, \quad x_h=\frac{h}{\sqrt{|\X|}}.
\]
Then $\varphi_f,x_h$ are vectors in the unit ball for any $f\in \F$, $h\in \H$, and we have
\[
L_{\X,f}(h)=\inP{\varphi_f}{x_h}+s_f.
\]
In particular we have
 \[
w_f(h)=\begin{cases}
    1 \quad& \inP{\varphi_f}{x_h}+s_f\leq \alpha,
    \\-1 \quad& \inP{\varphi_f}{x_h}+s_f\geq \alpha+\varepsilon,
    \\\star \quad& \text{otherwise}.
\end{cases}
\]
Hence, $\W$ can indeed be embedded into the class of affine classifiers  with margin $\varepsilon$. By \cite[Proposition 17]{alon2022theory} we have
\[
\VC(\W)\leq \frac{C}{\varepsilon^2}.
\]
Thus by \ref{Lem:Nets-Par}, there is some finite set $H\subset \H$, of size at most 
\[
|H|\leq \frac{C}{\varepsilon^2 \delta}\log \frac{e}{\delta},
\]
such that for any $f\in \F$, there is some $h\in H$, with $w_f(h)\neq -1$, that is \[
L_{\X,f}(h)\leq \alpha+ \varepsilon.
\] 
This implies the desired result.

\end{proof}

\begin{customthm}{\ref*{Thm:Par-Cover}}[Restated]\label{Proof:Par-Cover}
Let $\F\subseteq\{-1,+1,\star\}^{\X}$ be a partial binary class of VC dimension $d$ on a finite domain $\X$. For every $\varepsilon\in(0,1/2]$, there is an $\varepsilon$-disambiguation $\H\subseteq\{-1,+1\}^{\X}$ satisfying
\[
|\H|\leq(1/\varepsilon)^{cd\log(1/\varepsilon)},
\]
where $c>0$ is a universal constant.

This bound is optimal up to constant factors in the exponent: there are universal constants $c_0,\varepsilon_0>0$ such that, for every integer $d\geq1$ and every $\varepsilon\in(0,\varepsilon_0]$, some partial binary class of VC dimension $d$ has the property that every $\varepsilon$-disambiguation $\H$ satisfies
\[
|\H|\geq(1/\varepsilon)^{c_0d\log(1/\varepsilon)}.
\]
\end{customthm}
\begin{proof}[Proof of Theorem~\ref{Thm:Par-Cover}]
\emph{Upper bound.}
If $d=0$, no point receives both binary labels among the functions in $\F$, so a single total function extends the whole class. Assume henceforth that $d\geq1$, and put $\rho=\varepsilon/2$. Lemma~\ref{Lem:Par-Frac-cover} gives a distribution $\mu$ over total binary functions such that
\[
\Pr_{h\sim\mu}\bigl(L_{\X,f}(h)\leq\rho\bigr)\geq\delta
\quad\text{for every }f\in\F,
\qquad
\delta=\left(\frac{A}{\rho}\right)^{-Bd\log(e/\rho)},
\]
where $A>1$ and $B>0$ are universal constants. Apply Lemma~\ref{Lem:Par-Frac-to-Cover} with approximation error $\alpha=\rho$ and additional error $\rho$. This gives an $\varepsilon$-disambiguation of size
\[
|\H|\leq\frac{C_1}{\rho^2\delta}\log\frac{e}{\delta}
\leq(1/\varepsilon)^{cd\log(1/\varepsilon)}.
\]
The second inequality follows from $\log(1/\delta)=O(d\log^2(1/\varepsilon))$, using $\varepsilon\leq1/2$ and $d\geq1$ to absorb all remaining factors into the universal constant $c$.

\medskip
\noindent\emph{Lower bound.}
Fix $d\geq1$ and a sufficiently small $\varepsilon>0$, and set $n=\lfloor1/(4\varepsilon)\rfloor$. Combining Pabbaraju's construction \cite[Theorem~3 and Appendix~B]{pabbaraju2026optimal} with the graph-to-class reduction of \cite{alon2022theory} gives a partial binary class $\F_n$ on $n$ points of VC dimension one such that every exact disambiguation has size at least $n^{a\log n}$, for a universal $a>0$.
The construction gives such a class for every sufficiently large $n$.\footnote{In Pabbaraju's construction, the domain size $n$ progresses geometrically. Padding with undefined inputs extends the $n^{\Omega(\log n)}$ lower bound to every sufficiently large $n$.}

It remains to amplify this lower bound to VC dimension $d$, while allowing an $\varepsilon$-fraction of errors. 


We do this in three steps: pass to fractional disambiguations, take a direct product, and account for the allowed errors.

\smallskip
\noindent\emph{1. Fractional disambiguations and their dual.}
Write $h\succeq f$ if the total function $h$ agrees with the partial function $f$ on every defined input. A fractional exact disambiguation assigns nonnegative weights $w_h$ to total binary functions, subject to
\[
\sum_{h\succeq f}w_h\geq1
\qquad\text{for every }f\in\F_n.
\]
Let $q_n$ be the minimum total weight $\sum_h w_h$. We first show that $q_n$ is still large. Normalize optimal weights to a probability distribution. Each $f\in\F_n$ is extended by a draw from this distribution with probability at least $1/q_n$. Hence $k=\lceil q_n(\ln|\F_n|+1)\rceil$ independent draws fail to extend some member of $\F_n$ with probability at most $|\F_n|e^{-k/q_n}<1$. Some such sample is therefore an exact disambiguation. Since $q_n\geq1$ and $|\F_n|\leq3^n$, the one-block lower bound gives
\[
q_n\geq\frac{n^{a\log n}}{\ln|\F_n|+2}
\geq\frac{n^{a\log n}}{n\ln3+2}
\geq n^{(a/2)\log n}
\]
for all sufficiently large $n$. Thus, passing to the fractional relaxation loses only a constant factor in the exponent.

Linear-programming duality gives nonnegative weights $\lambda_f$ on the partial functions satisfying
\[
\sum_{f\in\F_n}\lambda_f=q_n,
\qquad
\sum_{f:\,h\succeq f}\lambda_f\leq1
\quad\text{for every total binary function }h.
\]
Consequently, $\nu_n(f)=\lambda_f/q_n$ is a probability distribution on $\F_n$ such that
\[
\Pr_{f\sim\nu_n}[h\succeq f]\leq\frac1{q_n}
\qquad\text{for every fixed total binary function }h.
\]
This is the form of the lower bound that we will use on each block.

\smallskip
\noindent\emph{2. Taking a direct product.}
Let $\G=\F_n^{\otimes d}$ be the class on $d$ disjoint blocks of $n$ points obtained by choosing a member of $\F_n$ on each block, with all choices allowed. Its VC dimension is exactly $d$: one point can be shattered in each block, while no shattered set can contain two points from the same block.

The fractional exact disambiguation number of $\G$ is exactly $q_n^d$. Indeed, taking products of the one-block weights $w_h$ gives a feasible fractional disambiguation of total weight $q_n^d$. Taking products of the dual weights $\lambda_f$ gives a feasible dual solution of the same value. In both cases the constraints factor across the blocks, proving the claimed equality.

\smallskip
\noindent\emph{3. Allowing approximation errors.}
Draw $f\in\G$ by choosing its restrictions $f_1,\ldots,f_d$ independently from $\nu_n$, and fix a total binary function $h$, with restrictions $h_1,\ldots,h_d$. Our choice $n=\lfloor1/(4\varepsilon)\rfloor$ ensures that an $\varepsilon$-approximation makes at most $\varepsilon dn\leq d/4$ disagreements in total. Every block on which $h_i\not\succeq f_i$ contributes at least one disagreement. Thus, $h_i\succeq f_i$ must hold on at least $\lceil d/2\rceil$ blocks.

For any specified set of $k$ blocks, independence and the bound from Step~1 give probability at most $q_n^{-k}$ of exact agreement on all of them. Taking $k=\lceil d/2\rceil$ and a union bound over the possible sets of blocks yields
\[
\Pr\bigl[h\text{ makes at most }\varepsilon dn\text{ disagreements with }f\bigr]
\leq\binom{d}{\lceil d/2\rceil}q_n^{-\lceil d/2\rceil}
\leq2^d q_n^{-d/2}.
\]
If $\H$ is an $\varepsilon$-disambiguation of $\G$, every sampled $f$ must be approximated by some $h\in\H$. A final union bound gives $1\leq|\H|\,2^d q_n^{-d/2}$, and hence
\[
|\H|\geq2^{-d}q_n^{d/2}
\geq2^{-d}n^{(a/4)d\log n}
\geq(1/\varepsilon)^{c_0d\log(1/\varepsilon)}.
\]
The last inequality holds for a universal $c_0>0$ and all sufficiently small $\varepsilon$, since $n=\Theta(1/\varepsilon)$. The required upper bound $\varepsilon_0$ on $\varepsilon$ is independent of $d$.
\end{proof}

\subsection{Covering Number Bounds }\label{Sec:Real-Covers-Proof}
Recall that
$H$ is an $(\alpha,\varepsilon)$-approximate cover of a function class $\F$, over finite domain $\X$, if for every $f\in \F$ there is some $h\in H$ such that
\[
|f(x)-h(x)|\leq \alpha
\]
for all but at most $\varepsilon|\X|$ points $x\in \X$.
\begin{lemma}\label{Lem:Cover-base}
{Let $\F$ be a class of real-valued functions on a finite domain $\X$ of size $n$. Let $\gamma>0$, $\varepsilon\in(0,1)$, and suppose that $d=\fat_\gamma(\F)<\infty$. Assume that, for some $\alpha\in[0,1]$ and $r>\gamma$, every $f\in\F$ satisfies $|f(x)|\leq r/2$ on all but at most $\alpha n$ points of $\X$.}
    
    Then $\F$ has an $(\frac{r+\gamma}{4},\alpha+\varepsilon)$-approximate cover $G$ of size 
    \[
   |G|\leq \left(\frac{C}{\varepsilon}\right)^{c\cdot d\log \frac{e}{\varepsilon}},
    \]
     where $ {C,c}>0$ are universal constants.
\end{lemma}
\begin{proof}
    For each $f\in \F$ define the partial function $h_f(x)$ to be $1$ if $f(x)\geq \frac{\gamma}{2}$, $-1$ if $f(x)\leq -\frac{\gamma}{2}$, and undefined if $f(x)\in (\frac{-\gamma}{2},\frac{\gamma}{2})$. Clearly the induced partial concept class $\H_\F=\{h_f\;:\; f\in \F\}$ has VC dimension at most $d$. Thus, applying Theorem~\ref{Thm:Par-Cover} with accuracy $\min\{\varepsilon,1/2\}$ gives some $H\subset\{-1,{+1}\}^\X$ of size at most \[
    |H|\leq \left(\frac{C}{\varepsilon}\right)^{c\cdot d\log \frac{e}{\varepsilon}},
    \]
    such that for each $h_f\in\H_\F$ there is $h\in H$, with ${L_{\X,h_f}(h)}\leq \varepsilon$. For each such $h$ define $\phi_h:\X\to[-r,r]$ by
  \[
    \phi_h(x)=\case{\frac{r-\gamma}{4}}{h(x)=1,}{\frac{\gamma-r}{4}}{h(x)=-1.}
    \]
{Fix $f\in\F$ and choose $h\in H$ with $L_{\X,h_f}(h)\leq\varepsilon$.} By assumption there are at most $\alpha|\X|$ points in $\X$, for which $|f(x)|>\frac{r}{2}$. At every other point, if either $h_f(x)=\star$ (so $f(x)\in [\frac{-\gamma}{2},\frac{\gamma}{2}]$) or $h_f(x)=h(x)$, then  
\[
|f(x)-\phi_h(x)|\leq \frac{r+\gamma}{4}.
\] 
For any $f\in\F$ there is some $h\in H$ such that $h_f(x)$ is defined and not equal $h(x)$ for no more than $\varepsilon n$ points. Thus there are at most $(\alpha+\varepsilon)|\X|$ points for which the above does not hold, and $\{\phi_h\;:\; h\in H\}$ is the desired $(\frac{r+\gamma}{4},\alpha+\varepsilon)$-approximate cover.

\end{proof}

\begin{lemma}\label{Lem:Cover-ind}
     Let $\F\subseteq[-1,+1]^\X$ be a function class over finite domain $\X$ of size $n$. Let {$\gamma>0$ and $\varepsilon\in(0,1)$} and assume that $d=\fat_\gamma(\F)$ is finite. Then there exists $H\subset [-1,+1]^\X$, an $(\frac{\gamma}{2}+\varepsilon,2\varepsilon\log \frac{e}{\varepsilon})$-approximate cover for $\F$ of size 
     \[
    |H|\leq \left(\frac{C}{\varepsilon}\right)^{c\cdot d\log^2 \frac{e}{\varepsilon}}. 
    \]
\end{lemma}

\begin{proof}

 We prove the following slightly more general statement by induction on
    $k$: for every $k\geq 0$, the class $\F$ has an
    \[
    \left(\frac{\gamma}{2}+2^{-k},k\varepsilon\right)
    \]
    approximate cover of size
    \[
    \left(\frac{C}{\varepsilon}\right)^{cdk\log \frac{e}{\varepsilon}}.
    \]
Choosing $k$ such that $\frac{1}{2^{k}}\le \varepsilon \le \frac{1}{2^{k-1}}$ then yields the desired bound.
    For $k=0$, the zero function gives a cover of
    radius $1$.

    {Assume the claim holds for $k$. If $k\varepsilon\geq1$, a single representative satisfies the claim for $k+1$, so assume $k\varepsilon<1$.} Let $H_k$ be the corresponding cover.
    Fix $\phi\in H_k$, and let $\F_\phi$ be the set of functions
    $f\in\F$ which are covered by $\phi$, that is, all functions $f\in \F$ satisfying
    \[
    |\{x\in \X: |f(x)-\phi(x)|>\gamma/2+2^{-k}\}|
    \leq
    k\varepsilon |\X|.
    \]
    Apply Lemma \ref{Lem:Cover-base} to the residual class
    \[
    \F_\phi-\phi=\{f-\phi:f\in\F_\phi\},
    \]
    with
    \[
    \frac{r}{2}=\frac{\gamma}{2}+2^{-k},
    \qquad
    \alpha=k\varepsilon.
    \]
    Translation by $\phi$ does not increase the $\gamma$-fat-shattering
    dimension, so the relevant dimension is still at most $d$.

    Lemma \ref{Lem:Cover-base} gives a cover of $\F_\phi-\phi$ with
    covering error $(k+1)\varepsilon$ and radius
    \[
    \frac{r+\gamma}{4}
    =
    \frac{\gamma}{2}+2^{-(k+1)}.
    \]
    Translating this cover back by adding $\phi$, clipping each function above $1$ and below $-1$, so it will be in $[-1,+1]^\X$, and doing this for every
    $\phi\in H_k$, gives the desired cover for the next step.  The size is
    multiplied by at most
    \[
    \left(\frac{C}{\varepsilon}\right)^{c\cdot d\log \frac{e}{\varepsilon}},
    \]
    which proves the induction.

\end{proof}

  
\begin{rtheorem}{Thm:ell_2-bound}\label{Proof:Cover-ell_2}
Let $n\geq1$, let $V\subseteq[-1,+1]^n$, let $\gamma>0$, and write $d=\fat_\gamma(V)<\infty$. For every $\varepsilon\in(0,1)$,
\[
\log N_2\left(V,\frac{\gamma}{2}+\varepsilon\right)
\leq C d\log^3\frac{e}{\varepsilon},
\]
where $C>0$ is a universal constant.
\end{rtheorem}
\begin{proof}[Proof of Theorem~\ref{Thm:ell_2-bound}]
View $V$ as a class of functions on $[n]$. By Lemma~\ref{Lem:Cover-ind}, for every $\alpha\in(0,1)$ it has a
\[
\left(\frac{\gamma}{2}+\alpha,2\alpha\log\frac{e}{\alpha}\right)
\]
approximate cover of size at most
\[
\left(\frac{C}{\alpha}\right)^{C d\log^2(e/\alpha)}.
\]
Clipping the representatives to $[-1,+1]^n$ cannot increase any coordinate error. The resulting cover is therefore a normalized $\ell_2$ cover at radius
\[
\sqrt{\left(\frac{\gamma}{2}+\alpha\right)^2+8\alpha\log\frac{e}{\alpha}}
\leq\frac{\gamma}{2}+4\sqrt{\alpha\log\frac{e}{\alpha}}.
\]
Choose $\alpha=c\varepsilon^4$, where $c>0$ is a sufficiently small universal constant. Then $4\sqrt{\alpha\log(e/\alpha)}\leq\varepsilon$, and $\log(e/\alpha)=O(\log(e/\varepsilon))$. Taking the logarithm of the cover size gives
\[
\log N_2\left(V,\frac{\gamma}{2}+\varepsilon\right)
\leq C' d\log^3\frac{e}{\varepsilon},
\]
as required.
\end{proof}

\subsection{Main Result }\label{Sec:Main-Result-Proof}

The characterization of \cite{aiyer2026scale} states that $\gamma$-uniform convergence holds exactly when $\fat_{\gamma'}(\F)<\infty$ for every $\gamma'>\gamma$. Our theorem gives a sharp quantitative guarantee from the stronger assumption $d=\fat_\gamma(\F)<\infty$.

For $S=(x_1,\ldots,x_n)$, write
\[
V_S=\{(f(x_1),\ldots,f(x_n)):f\in\F\}.
\]
Then $\fat_\gamma(V_S)\leq\fat_\gamma(\F)$, so the vector bounds from the preceding sections apply to every sample.

Define the positive and negative representativeness  
\[
\Rep_{\D}^+(\F,S)
=
\sup_{f\in\F}
\Ex_{x\sim\D} f(x)
-
\frac{1}{n}\sum_{i=1}^n f(x_i)
,  \quad \Rep_{\D}^-(\F,S)
=
\sup_{f\in\F}
\frac{1}{n}\sum_{i=1}^n f(x_i)
-
\Ex_{x\sim\D} f(x).
\]
Note that $\Rep^+$ is the notion of representativeness defined in \cite{understandingml}. Its advantage is that it is consistent with our definition of Rademacher complexity by having no absolute value inside the expectation, so we can relate it to the Rademacher complexity.

The following  concentration inequality is known as  McDiarmid's inequality 

\begin{theorem}[ McDiarmid's inequality \cite{mcdiarmid1989method}]\label{Thm:Mcdim}
    Say that $S,S'\in \X^n$ are $i$-neighbors if $S_j=S'_j$ for all $j\neq i$. Let $f:\X^n\to \R$ be some function such that for every pair of $i$-neighbors $S,S'\in \X^n$, we have \[
    |f(S)-f(S')|\leq c_i,
    \]
    where $c_1,c_2\dots c_n$, are some fixed positive constants. Then, for every $\varepsilon>0$,
    \[
    \Pr_{S\sim \D^n}
    \left[
    \left|f(S)-\Ex_{S'\sim \D^n} f(S')\right|
    \geq \varepsilon
    \right]
    \leq
    2\exp\left(\frac{-2\varepsilon^2}{\sum_{i=1}^n c_i^2}\right).
    \]
    In particular, if $f(S)$ is one of ${\Rep_\D}(\F,S)$, ${\Rep_\D}^+(\F,S)$, ${\Rep_\D}^-(\F,S)$, we may take $c_i=2/n$ for all $1\leq i\leq n$, and get 
     \[
    \Pr_{S\sim \D^n}
    \left[
    \left|f(S)-\Ex_{S'\sim \D^n} f(S')\right|
    \geq \varepsilon
    \right]
    \leq
    2\exp\left(\frac{-\varepsilon^2n}{2}\right).
    \]
\end{theorem}

\begin{theorem}\label{Thm:sharp-bound}
 Let \(\F\subset[-1,+1]^{\mathcal X}\), let \(\gamma\in (0,1)\), and assume that 
$d=\fat_\gamma(\mathcal F)<\infty $ is finite.
Then there is a universal constant \(C>0\) such that for every \(n\ge 1\) and every $S\in\X^n$,
\[
\Rad(V_S)
\le
\frac{\gamma}{2}
+C\sqrt{\frac dn}
\]
{Consequently, for every $\varepsilon,\delta\in(0,1)$, if}
\[
n
\ge
C\frac{
d+\log(1/\delta)
}{\varepsilon^2},
\]
then for every distribution $\D$ over $\X$, with probability at least \(1-\delta\) over \(S\sim\mathcal D^n\),
\[
\Rep_{\mathcal D}(\mathcal F,S)\le \gamma+\varepsilon.
\]
\end{theorem}

\begin{proof}
Fix $S\in\X^n$. Apply Theorem~\ref{Thm:Chain-Imp} to $V_S$ with $\alpha=\gamma/2$ to obtain
\[
\Rad(V_S)\leq \frac{\gamma}{2}+ \frac{32}{\sqrt{n}}\int_0^1\sqrt{\log N_2(V_S,\frac{\gamma}{2}+t^2)}dt.
\]

By Theorem~\ref{Thm:ell_2-bound} and $\fat_\gamma(V_S)\leq d$, for every $t\in(0,1)$,
\[
{\log N_2(V_S,\frac{\gamma}{2}+t^2)}
\le
C\cdot d\cdot \log^3\frac {e}{t},
\]
where $C>0$ is some universal constant.
Hence the integral is bounded \[
\int_0^1\sqrt{\log N_2(V_S,\frac{\gamma}{2}+t^2)}dt\leq \int_0^1\sqrt{C\cdot d\cdot \log^3\frac {e}{t}}dt\leq C'\sqrt{d},
\]
where in the last inequality we used convergence of the integral of $\log^{3/2}\frac {e}{t}$ over $[0,1]$.
Thus we indeed have
\[
\Rad(V_S)
\le
\frac{\gamma}{2}
+
C\sqrt{\frac{d}{n}}.
\]

The bound holds for every $S$. By symmetrization \cite[Lemma~26.2]{understandingml}, we have
\[
\Ex_{S\sim\mathcal D^n}\Rep_{\mathcal D}^+(\mathcal F,S)
\le
2\Ex_{S\sim\mathcal D^n}\Rad(V_S),
\]
so, after changing the universal constant,
\[
\Ex_{S\sim\mathcal D^n}\Rep_{\mathcal D}^+(\mathcal F,S)
\le
\gamma
+
C\sqrt{\frac dn}.
\]

Apply the previous bound with 
\[
n
\ge
C\frac{
d
}{\varepsilon^2},
\]
to get

\[
\Ex_{S\sim\mathcal D^n}\Rep_{\mathcal D}^+(\mathcal F,S)
\le  \gamma+C\sqrt{\frac{d}{n}}\leq
\gamma+\frac{\varepsilon}{2}.
\]
Using McDiarmid’s inequality \ref{Thm:Mcdim}, yields
\[
\Pr_{S\sim\mathcal D^n}
\left[
\Rep_{\mathcal D}^+(\mathcal F,S)>\gamma+\varepsilon
\right]\leq \Pr_{S\sim\mathcal D^n}
\left[
\Rep_{\mathcal D}^+(\mathcal F,S)>\Ex_{S'\sim\mathcal D^n}\Rep^+_{\mathcal D}(\mathcal F,S')+\frac{\varepsilon}{2}
\right]
\le
2\exp(-\frac{\varepsilon^2 n}{8}).
\]
Hence if we also have
\[
n\geq C\frac{\log 1/\delta}{\varepsilon^2},
\]
the above probability is at most \(\delta/2\). By replacing $\F$ with $-\F$, we get the same bound for $\Rep_{\mathcal D}^-(\mathcal F,S)$, and since
\[
\Rep_{\mathcal D}(\mathcal F,S)=\max \Big(\Rep_{\mathcal D}^-(\mathcal F,S),\Rep_{\mathcal D}^+(\mathcal F,S)\Big),
\]
we may use union bound to deduce the desired result.
\end{proof}

{We finish with the matching lower bound.}

\begin{prop}\label{prop:lower-bnd}
There are universal constants $c,\varepsilon_0,\delta_0>0$ such that, for every $\gamma\in(0,1)$ and integer $d\geq1$, there is a class $\F\subseteq[-1,+1]^\X$ with $\fat_\gamma(\F)=d$ satisfying the following. For every sufficiently small $\eps,\delta>0$ , and
\[
 n\leq c\frac{d+\log(1/\delta)}{\varepsilon^2},
\]
there is a distribution $\D$ such that
\[
 \Pr_{S\sim\D^n}\bigl[\Rep_\D(\F,S)>\gamma+\varepsilon\bigr]>\delta.
\]
\end{prop}
\begin{proof}
Let $Y=[d]\times\{-1,+1\}$ and
\[
 \H=\{h_a:a\in\{-1,+1\}^d\},
 \qquad h_a(j,y)=a_jy.
\]
This symmetric class has $2^d$ functions and VC dimension $d$. Its uniform-convergence lower bound follows from the standard agnostic-learning lower bound for all binary predictors on $[d]$: their zero-one losses on $Y$ are $(1-h_a)/2$; see \cite[Theorem~6.8 and Chapter~28]{understandingml}. Thus, after adjusting universal constants, for the stated range of $\varepsilon,\delta,n$ there is a distribution $\D_2$ on $Y$ such that
\[
 \Pr_{K\sim\D_2^n}\bigl[\Rep_{\D_2}(\H,K)>2\varepsilon\bigr]>\delta.
\]

Set $\X=[0,1]\times Y$, let $\B$ be the measurable functions from $[0,1]$ to $(-1,+1)$, and define
\[
 \F=\{f_{h,b}:h\in\H,\ b\in\B\},
 \qquad
 f_{h,b}(t,z)=\left(1-\frac{\gamma}{2}\right)h(z)+\frac{\gamma}{2}b(t).
\]
For a fixed $h$, any two functions $f_{h,b}$ and $f_{h,b'}$ differ by strictly less than $\gamma$ at every point. Consequently, two different labelings of a $\gamma$-fat-shattered set cannot be realized using the same $h$. A shattered set of size $k$ therefore requires $2^k\leq|\H|=2^d$, proving $\fat_\gamma(\F)\leq d$. Conversely, take $b=0$ and the points $(0,(j,+1))$, for $j\in[d]$. These are $\gamma$-fat-shattered with thresholds zero, since $2-\gamma>\gamma$. Hence $\fat_\gamma(\F)=d$.

Let $\D_1$ be uniform on $[0,1]$ and take $\D=\D_1\times\D_2$. For a sample $S=((t_i,z_i))_{i=1}^n$, write $T=(t_i)_{i=1}^n$ and $K=(z_i)_{i=1}^n$. The symmetry of $\H$ and $\B$, and the independent choices of $h$ and $b$ in the supremum, give
\[
 \Rep_\D(\F,S)
 =\left(1-\frac{\gamma}{2}\right)\Rep_{\D_2}(\H,K)
 +\frac{\gamma}{2}\Rep_{\D_1}(\B,T).
\]
For every finite $T$, the supremum defining $\Rep_{\D_1}(\B,T)$ is $2$: choose $b$ arbitrarily close to $-1$ on the sampled points and arbitrarily close to $+1$ elsewhere. Thus
\[
 \Rep_\D(\F,S)
 =\gamma+\left(1-\frac{\gamma}{2}\right)\Rep_{\D_2}(\H,K).
\]
Since $1-\gamma/2>1/2$, the lower bound for $\H$ proves the proposition.
\end{proof}

\begin{rtheorem}{Thm:Main result}\label{Proof:Main result}

Let $\F\subseteq[-1,+1]^\X$, let $\gamma\in(0,1)$, and suppose that $d:=\fat_\gamma(\F)<\infty$. Then $\F$ satisfies $\gamma$-uniform convergence, with
\[
 m_\gamma(\varepsilon,\delta)
 =O\left(\frac{d+\log(1/\delta)}{\varepsilon^2}\right)
 \qquad \varepsilon,\delta\in(0,1).
\]
Moreover, for every $\gamma\in(0,1)$ and integer $d\geq1$, there is a class $\F$ with $\fat_\gamma(\F)=d$ such that, for all sufficiently small $\varepsilon,\delta>0$,
\[
 m_\gamma(\varepsilon,\delta)
 =\Omega\left(\frac{d+\log(1/\delta)}{\varepsilon^2}\right).
\]
\end{rtheorem}
\begin{proof}[Proof of Theorem \ref{Thm:Main result}]
    {The upper bound follows from Theorem~\ref{Thm:sharp-bound}, and the lower bound follows from Proposition~\ref{prop:lower-bnd}.}
\end{proof}

\paragraph{Use of AI tools.}
The authors used ChatGPT to assist with writing, reviewing proofs, and developing
parts of several arguments. Specifically it suggested applying  $\varepsilon$-nets result to a fractional cover,  provided Theorem \ref{Thm:Net-Hilbert-Halfspaces} and Lemma \ref{Lem:Geo-center}, and helped optimize several of the technical results and lower bounds.
The authors revised and verified the
arguments presented in the paper and take responsibility for its contents.

\bibliographystyle{alpha}
\bibliography{bib}

\end{document}